\documentclass[10pt,twocolumn,letterpaper]{article}

\usepackage[pagenumbers]{iccv} 

\definecolor{iccvblue}{rgb}{0.21,0.49,0.74}
\usepackage[pagebackref,breaklinks,colorlinks,allcolors=iccvblue]{hyperref}

\newcommand*{\affaddr}[1]{#1} 
\newcommand*{\affmark}[1][*]{\textsuperscript{#1}}
\newcommand*{\email}[1]{\texttt{#1}}

\def\paperID{1079} 
\def\confName{ICCV}
\def\confYear{2025}

\title{Beyond Generation: Unlocking Universal Editing via Self-Supervised Fine-Tuning}

\author{%
Harold Haodong Chen\affmark[1,2]~~~~~ 
Harry Yang\affmark[1,2]~~~~~
Ser-Nam Lim\affmark[1,3]~~~~~\\
\affaddr{\affmark[1]Everlyn AI~~~~~~~}
\affaddr{\affmark[2]HKUST~~~~~~~}
\affaddr{\affmark[3]UCF~~~~~~~}\\
\small\textbf{\email{Project page:\textcolor{magenta}{\url{https://haroldchen19.github.io/UES-Page/}}}}
}

\begin{document}

\twocolumn[{
\renewcommand\twocolumn[1][]{#1}
\maketitle
\begin{center}
    \vspace{-10pt}
   \includegraphics[width=1\linewidth]{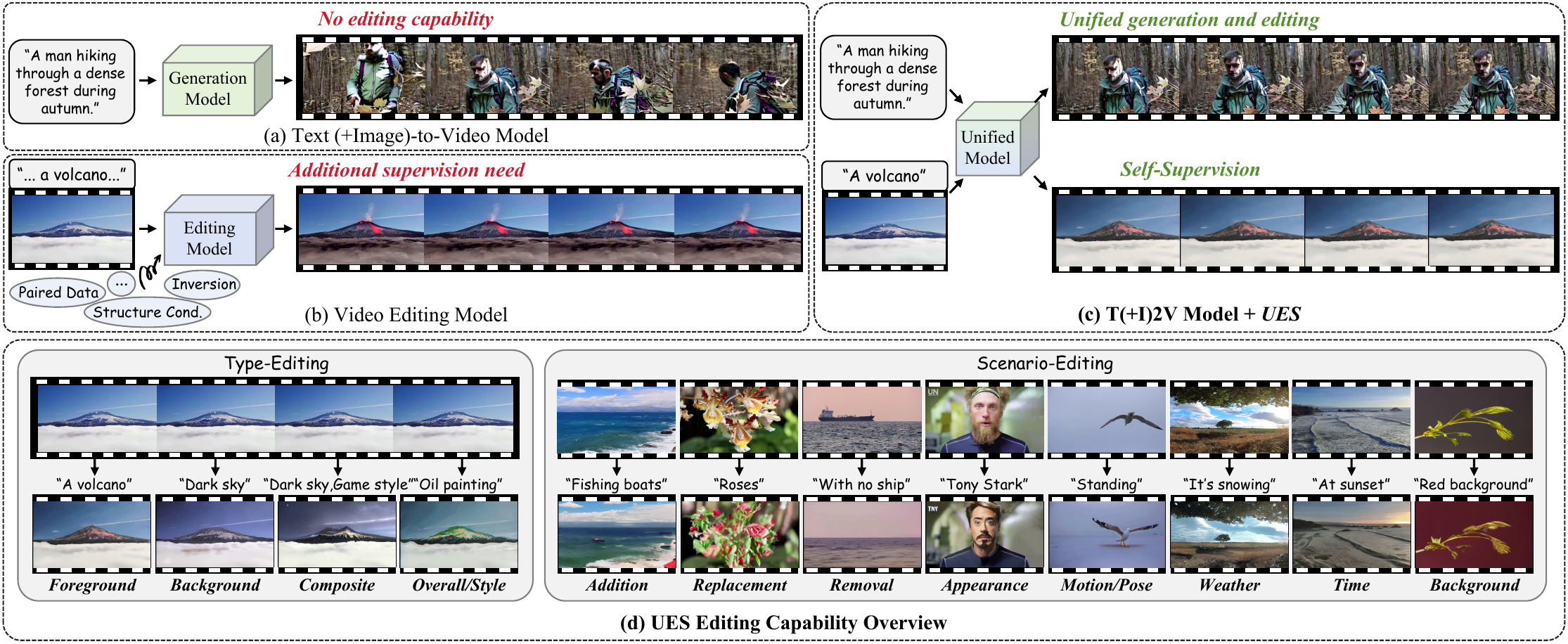}
   \vspace{-1.8em}
    \captionof{figure}{\label{fig:example}
    \textbf{\texttt{UES}: Unlocking Universal Editing via Self-Supervision.} We introduce UES, a lightweight self-supervised fine-tuning strategy that enables text(+image)-to-video models to achieve universal editing capabilities without relying on additional supervision. (\textbf{\textit{a}}) Current T(+I)2V models excel at generation but lack inherent editing capabilities and incur high computational costs for fine-tuning. (\textbf{\textit{b}}) Existing video editing models often require additional supervision. (\textbf{\textit{c}}) By applying UES, a unified model is achieved, seamlessly combining generation and editing functionalities. (\textbf{\textit{d}}) Compared to existing editing models, which are limited to four editing types, UES extends editing capabilities across four types and eight diverse editing scenarios.}
\end{center}
}]

\begin{abstract}
Recent advances in video generation have outpaced progress in video editing, which remains constrained by several limiting factors, namely: (a) the task's dependency on supervision severely 
limits generality, (b) an unnecessary artificial separation between the generation and editing task, and (c) the high computational costs of training a video model. In this work, we propose \textbf{\texttt{UES}} (\textit{Unlocking Universal Editing via Self-Supervision}), a lightweight self-supervised fine-tuning strategy that transforms generation models into \textbf{unified generation-editing systems} through self-supervised semantic alignment. Our approach establishes a dual-conditioning mechanism where original video-text pairs jointly provide \textit{visual} and \textit{textual semantics}, enabling structured learning of intrinsic spatiotemporal correspondences. Key advantages include: (i) \textbf{Universality} through supervision-free adaptation to diverse editing tasks, (ii) \textbf{Unification} of generation and editing applicable to most text(+image)-to-video model, and (iii) \textbf{Efficiency} via lightweight fine-tune that reduces tunable parameters by \textbf{92.67\%}. To enable systematic evaluation, we introduce \textbf{\texttt{OmniBench-99}}, a comprehensive benchmark spanning 99 videos across humans/animals, environments, and objects, comprising 4 editing types and 8 scenarios. Extensive experiments show UES enables models without inherent editing capability to perform powerful and universal editing while preserving or even enhancing their original generation performance. 

\end{abstract}    
\vspace{-1em}
\section{Introduction}
\label{sec:intro}

\vspace{-0.6em}
Recent advances in generative artificial intelligence have driven remarkable progress in video generation \cite{videoworldsimulators2024, yin2024causvid, li2025minimax}, achieving high-quality, complex, and longer videos. These developments are fueled by large-scale high-quality datasets and substantial computational resources. As a complementary task, video editing, 
a form of controllable video generation, has gained attention for its applications in content creation \cite{mou2024revideo, fang2024vivid, geyer2023tokenflow}. 

Despite its potential, existing video editing methods face several critical challenges as shown in Fig.~\ref{fig:example}(a, b). \textbf{First}, most approaches rely heavily on additional supervision signals, such as structural conditions \cite{xing2024make, liang2024flowvid, wang2024videocomposer, feng2024ccedit, bar2022text2live, yan2023motion, zhao2023make, couairon2023videdit, lee2023shape, khachatryan2023text2video}, DDIM inversion \cite{liew2023magicedit, couairon2023videdit, wu2023tune, ku2024anyv2v, shin2024edit, geyer2023tokenflow, jeong2023ground, zhao2023controlvideo, liu2024video, tu2024motioneditor}, or paired video data \cite{cheng2023consistent, qin2023instructvid2vid}. This dependency not only increases the complexity of these methods but also limits their 
generality \cite{sun2024diffusion}. \textbf{Second}, unlike the image domain, where unified models for generation and editing have been extensively explored \cite{xia2024dreamomni, chen2024unireal, xiao2024omnigen}, such unified frameworks are rare in the video domain. Current video models typically separate generation and editing into distinct processes, resulting in fragmented solutions that lack flexibility and scalability. Specifically, for practical purposes, this would mean a cumbersome procedure of deploying and maintaining separate models for generation and editing. \textbf{Third}, the computational cost of achieving such a unified framework is prohibitively high, particularly when relying on additional supervision signals and large-scale training \cite{zhang2024vast, videoworldsimulators2024, li2025minimax}. This renders existing methods less practical for real-world applications, where computational efficiency is a key requirement.

To address these challenges, we propose \textbf{\texttt{UES}} (\textit{Unlocking Universal Editing via Self-Supervision}), a novel lightweight fine-tuning strategy designed to strengthen video generation models while enabling universal video editing capabilities (see Fig.~\ref{fig:example}(c)). UES is built on the principle of learning the intrinsic semantic correspondence between original text-video pairs. In this setup, UES introduces the original video as an additional denoising condition alongside the text caption, forming a dual-conditioning mechanism. The \textit{textual semantics} provided by the text together with the \textit{visual semantics} delivered by the video encourage the model 
to associate these semantics in a structured and meaningful manner. Once fine-tuned, UES is capable of controlling the generation based on the condition video, in effect obtaining an innate video editing capability by using a textual description (delta prompt, \textit{e.g.}, ``a volcano") of the desired edits to the video.
Specifically, UES features:
\begin{itemize}[leftmargin=*]
    \item[$\blacktriangleright$] \textbf{Universality:} Eliminates reliance on additional supervision signals, adapting to diverse editing tasks through self-supervised learning.
    \item[$\blacktriangleright$] \textbf{Unification:} Achieves a unified framework with minimal modifications, seamlessly integrating video generation and editing while balancing quality and flexibility.
    \item[$\blacktriangleright$] \textbf{Efficiency:} Requires only lightweight fine-tuning, significantly reducing tunable parameters by $92.67\%$.
\end{itemize}

To comprehensively evaluate UES’s universal editing capabilities, we further introduce \textbf{\texttt{OmniBench-99}}, a comprehensive benchmark. Unlike existing generative video editing benchmarks \cite{feng2024ccedit, wu2023cvpr, perazzi2016benchmark}, which primarily focus on a limited set of editing types (see Fig.~\ref{fig:example}(d)), OmniBench-99, including 99 diverse videos across three categories (humans/animals, environments, and objects), not only assesses four editing types but also uniquely evaluates eight editing scenarios, thereby setting a new standard for video editing research.

To summarize, this work contributes in threefold:
\begin{itemize}[leftmargin=*]
    \item We propose \textbf{\texttt{UES}}, a novel lightweight self-supervised fine-tuning strategy that unlocks generation models' universal video editing capability. To our knowledge, UES represents a pioneering effort toward achieving a unified framework for video generation and editing.
    \item We explore adopting the same video as the denoising target while conditioning on itself for the first time, effectively learning text-video semantic correspondence while preserving diffusion models' self-supervised paradigm. This enables more flexible unified generation and editing.
    \item We construct and release a comprehensive high-quality annotated video editing evaluation benchmark, which can serve as a valuable resource for follow-up research.
    \item Extensive experiments demonstrate the efficiency and versatility of UES in enabling advanced video generation models (\textit{i.e.}, DynamiCrafter \cite{xing2023dynamicrafter} and VideoCrafter2 \cite{chen2024videocrafter2}), highlighting its potential to push the boundaries of controllable video generation.
\end{itemize}

\section{Related Work}
\label{sec:related}

\makeatletter
\renewcommand\subsubsection{\@startsection{subsubsection}{3}{\z@}%
                                     {-3.25ex\@plus -1ex \@minus .2ex}%
                                     {-1em}%
                                     {\normalfont\normalsize\bfseries}}
\makeatother

\vspace{-0.6em}
Recent progress in diffusion models (DMs) \cite{ho2020denoising, sohl2015deep} has enabled impressive text-to-image (T2I) generation \cite{nichol2021glide, ramesh2022hierarchical, rombach2022high, saharia2022photorealistic}, with extensions to video through video diffusion models (VDMs) \cite{ho2022video} using space-time factorized U-Net \cite{ronneberger2015u}. Modern text-to-video (T2V) models, based on U-Net \cite{chen2024videocrafter2, guo2023animatediff}, DiT \cite{kong2024hunyuanvideo, zhang2024vast, lin2024open, zheng2024open}, and autoregression \cite{deng2024autoregressive, hong2022cogvideo, wang2024emu3}, achieve high-quality, complex video generation but depend heavily on large datasets and high computational costs, limiting scalability.

Video editing adapts video generation for precise modifications. Drawing inspiration from image editing \cite{brooks2023instructpix2pix, couairon2022diffedit, meng2021sdedit}, several studies \cite{yang2023rerender, bar2022text2live, lee2023shape, khachatryan2023text2video, chai2023stablevideo, feng2024ccedit, wu2024fairy} employ image-to-image (I2I) models to achieve frame-based editing. The introduction of one-video tuning by TAV \cite{wu2023tune} has prompted research \cite{ceylan2023pix2video, zhang2024towards, ouyang2024codef, jeong2024vmc, liu2024video, zhao2023motiondirector} into fine-tuning T2V models for specific video structures, although this may limit usability and practical applications. To improve precision, external controls such as structural conditions (\textit{e.g.}, poses \cite{ma2024follow, zhao2023controlvideo, tu2024motioneditor}, depth maps \cite{xing2024make, kara2024rave, wang2024videocomposer}), attention features \cite{qi2023fatezero, yang2024fresco, li2024vidtome, wu2024fairy}, or DDIM inversion guidance \cite{wu2023tune, xing2024simda, qi2023fatezero, ceylan2023pix2video, ling2024motionclone} are often introduced. While these methods offer promising approaches to video editing, their reliance on additional controls can limit the model’s versatility (\textbf{Appendix}~\S\ref{app_overview}). Additionally, the absence of unified frameworks for video generation and editing has led to fragmented solutions. In this work, we propose UES, a lightweight fine-tuning strategy that unifies video generation and editing within a self-supervised framework. UES eliminates the need for external supervision, reduces computational overhead, and enables fine-grained, versatile editing, advancing the goal of universal video modeling.

\begin{figure}[t]
    \centering
    \vspace{-0.4em}
   \includegraphics[width=1\linewidth]{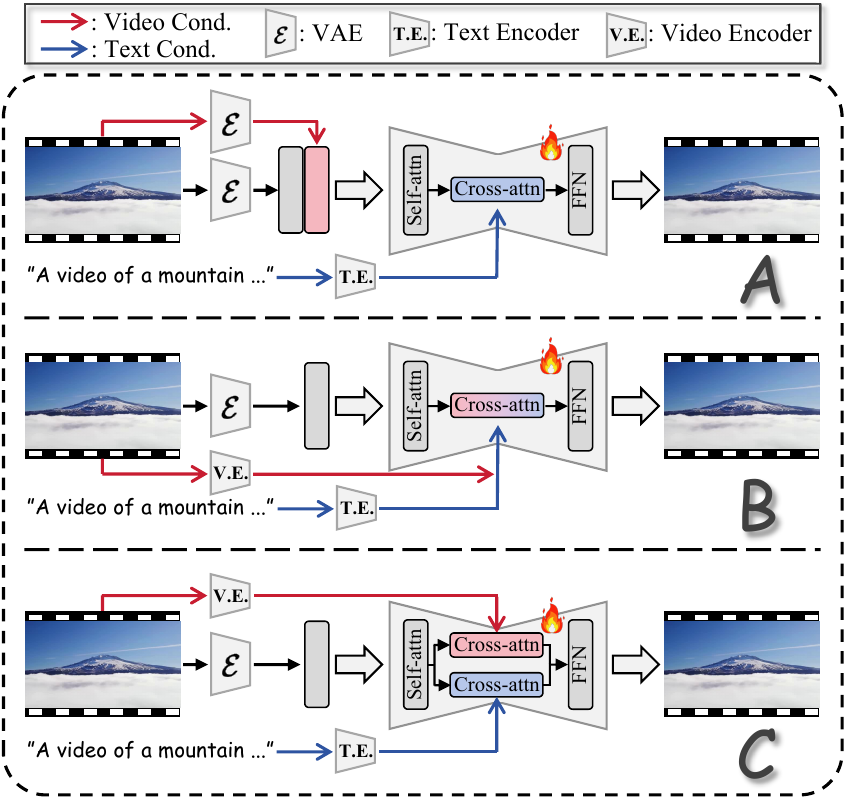}
   \vspace{-1.8em}
    \captionof{figure}{\label{fig:video_cond}
    Potential structures of UES video condition encoding. }
    \vspace{-1.2em}
\end{figure}

\section{Preliminary: Video Diffusion Models}
\vspace{-0.4em}
\noindent\textbf{Diffusion Models (DMs)}~\cite{ho2020denoising, sohl2015deep} are probabilistic models that gradually corrupt data $\mathbf{x}_0\sim p_{\mathrm{data}}(\mathbf{x})$ into Gaussian noises $\mathbf{x}_T\sim\mathcal{N}(0,\mathbf{I})$ through a forward process, and learn to reverse this process by denoising. The forward process $q(\mathbf{x}_t|\mathbf{x}_0,t)$, defined over $T$ timesteps, progressively adds noise to the data sample $\mathbf{x}_0$ to yield $\mathbf{x}_t$ through a parameterization trick. The denoising process $p_\theta(\mathbf{x}_{t-1}|\mathbf{x}_t,t)$ denoises the noisy input $x_{t}$ to obtain a cleaner data $x_{t-1}$ through a denoising network $\epsilon_\theta\left(\mathbf{x}_t,t\right)$, which is supervised by:
\setlength\abovedisplayskip{6pt}
\setlength\belowdisplayskip{4.4pt}
\begin{equation}
    \min_\theta\mathbb{E}_{t,\mathbf{x}\sim p_\text{data},\epsilon\sim\mathcal{N}(0,\mathbf{I})}\|\epsilon-\epsilon_\theta\left(\mathbf{x}_t; \mathbf{c},t\right)\|_2^2, \label{eq_dm}
\end{equation}

\noindent where $\epsilon$ refers to the sampled ground truth noise, $\theta$ represents the learnable parameters, and $\mathbf{c}$ indicates possible conditions. Once trained, the model generates denoised data $\mathbf{x}_0$ by iteratively denoising a random noise $\mathbf{x}_T$.

\noindent\textbf{Latent Diffusion Models (LDMs)}~\cite{rombach2022high, ho2022imagen} improve the computational efficiency by working in a learned compact latent space instead of the original pixel space. 
For a given video $\mathbf{x}\in\mathbb{R}^{L\times3\times H\times W}$, we first encode it into a latent representation $\mathbf{z}=\mathcal{E}(\mathbf{x})$ via an autoencoder $\mathcal{E}$, where $\mathbf{z}\in\mathbb{R}^{L\times C\times h\times w}$. The diffusion process $\mathbf{z}_{t}=p(\mathbf{z}_{0},t)$ and denoising process $\mathbf{z}_t=p_\theta(\mathbf{z}_{t-1},\mathbf{c},t)$ are then performed in the latent space. Finally, the generated videos are obtained through the decoder $\hat{\mathbf{x}}=\mathcal{D}(\mathbf{z})$.

\vspace{-0.4em}
\section{\texttt{UES}: Universal Editing via Self-Supervision}
\label{sec:method}

\vspace{-0.6em}
The primary objective of our work is to achieve unified video generation and editing. While previous methods have advanced controlled editing in specific scenarios by introducing additional controls, generalizable solutions remain challenging. To address this, UES introduces the original video as an additional condition alongside the text caption, forming a dual-conditioning mechanism that learns the intrinsic semantic correspondence between these modalities in a self-supervised manner.

\begin{figure}[t]
    \centering
    \vspace{-0.4em}
   \includegraphics[width=1\linewidth]{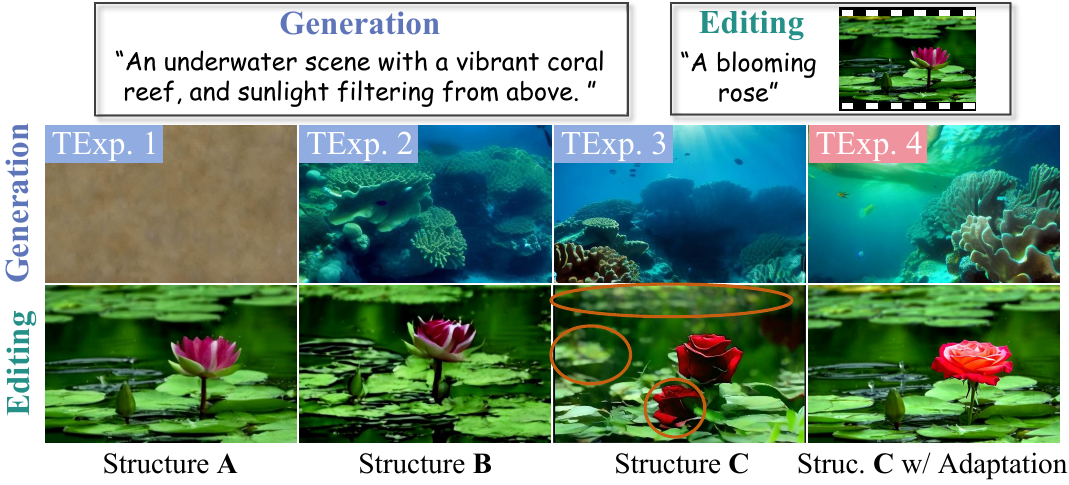}
   \vspace{-1.8em}
    \captionof{figure}{\label{fig:toy_exp}
    Toy experiments of the generation and editing capability of three structures in Fig.~\ref{fig:video_cond}. \textbf{TExp.~1} (Stuc.~A): Over-relies on video conditions, neglecting text input. \textbf{TExp.~2} (Stuc.~B): Combines conditions but causes semantic ambiguity. \textbf{TExp.~3} (Struc.~C): Dual-path strategy enables both generation and editing. \textbf{TExp.~4} (Struc.~C w/ Adaptation): Adaptation (Fig.~\ref{fig:ues_cond}) further enhances spatiotemporal learning, \textit{e.g.}, \textbf{preserving unedited areas}.}
    \vspace{-1.2em}
\end{figure}

\vspace{-0.3em}
\subsection{Video Condition Encoding}
\vspace{-0.4em}
To effectively encode video information into the denoising process, we explore three potential structures (Struc.) for video condition encoding, as illustrated in Fig.~\ref{fig:video_cond}, along with several toy experiments (TExp.) shown in Fig.~\ref{fig:toy_exp}.

\vspace{-1.3em}
\subsubsection*{Structure A: Concatenation of Video Latents.}
Struc.~\textcolor{magenta}{A} represents a straightforward approach to encoding the video condition by concatenating with the input video latent along the channel dimension before being fed into the denoising network. However, since the video condition is derived from the input video itself, this approach causes the model to focus excessively on reconstructing the input video from the condition (TExp.~\textcolor{magenta}{1}), as well as sacrificing its original generation capabilities.

\vspace{-1.4em}
\subsubsection*{Structure B: Joint Conditioning.}
Struc.~\textcolor{magenta}{B} improves upon Struc.~\textcolor{magenta}{A} by introducing the video condition processed alongside the text condition to form a combined representation, $\mathbf{E}_\mathrm{combine}$, which interacts with the intermediate features, $\mathbf{E}_\mathrm{in}$, via cross-attention within the denoising network:
\begin{equation}
    \mathbf{E}_\mathrm{out}=\text{Softmax}(\frac{\mathbf{QK}_\mathrm{combine}^{\top}}{\sqrt{d}})\mathbf{V}_\mathrm{combine},
\end{equation}
where $\mathbf{Q}=\mathbf{E}_\mathrm{in}\mathbf{W}_\mathbf{Q}$, $\mathbf{K}_\mathrm{combine}=\mathbf{E}_\mathrm{combine}\mathbf{W}_\mathbf{K}$, and $\mathbf{V}_\mathrm{combine}=\mathbf{E}_\mathrm{combine}\mathbf{W}_\mathbf{V}$ accordingly. Here, $\mathbf{W}_\mathbf{Q}$, $\mathbf{W}_\mathbf{K}$, and $\mathbf{W}_\mathbf{V}$ are learnable projection matrices.
This design enables simultaneous conditioning on video and text, maintaining generation capability when only text is provided. However, due to the high similarity between the video condition and the denoising target, the model tends to prioritize video features over textual guidance, resulting in insufficient editing capability, as demonstrated in TExp.~\textcolor{magenta}{2}.

\begin{figure}[t]
    \centering
    \vspace{-0.4em}
   \includegraphics[width=1\linewidth]{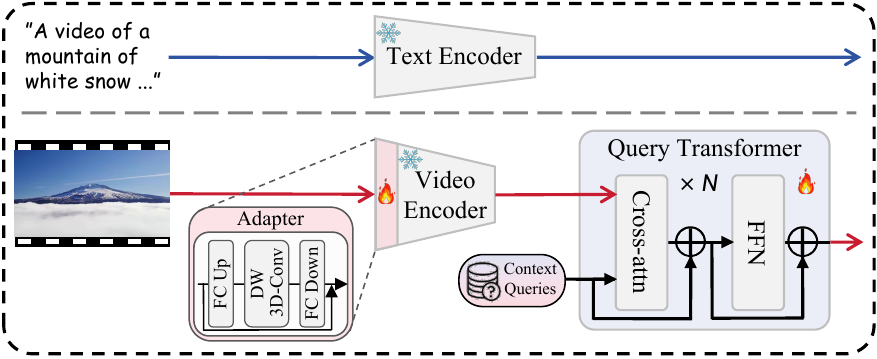}
   \vspace{-1.8em}
    \captionof{figure}{\label{fig:ues_cond}
    Illustration of UES condition modeling.}
    \vspace{-1.2em}
\end{figure}

\vspace{-1.34em}
\subsubsection*{Structure C: Dual-Path Conditioning.}
To address the limitations of the above two structures, we further explore Struc.~\textcolor{magenta}{C}, which employs dual-path cross-attention mechanisms for the video and text conditions representations, $\mathbf{E}_\mathrm{vid}$ and $\mathbf{E}_\mathrm{txt}$:
\setlength\abovedisplayskip{3.6pt}
\setlength\belowdisplayskip{3.6pt}
\begin{equation}
    \text{Softmax}(\frac{\mathbf{QK}_\mathrm{txt}^{\top}}{\sqrt{d}})\mathbf{V}_\mathrm{txt}+\text{Softmax}(\frac{\mathbf{QK}_\mathrm{vid}^{\top}}{\sqrt{d}})\mathbf{V}_\mathrm{vid},
\end{equation}
where the queries $\mathbf{Q}$ derive from an intermediate representation $\mathbf{E}_\mathrm{lat}$ via $\mathbf{Q}=\mathbf{W}_\mathbf{Q}^{(i)}\cdot \mathbf{E}_\mathrm{lat}$. The keys and values for text and video conditions are calculated as follows: 
\begin{equation}
    \left\{
    \begin{aligned}
    \mathbf{K}_\mathrm{txt}=\mathbf{W}_\mathbf{K}^{(i)}\cdot \mathbf{E}_\mathrm{txt}, \mathbf{V}_\mathrm{txt}=\mathbf{W}_\mathbf{V}^{(i)}\cdot \mathbf{E}_\mathrm{txt}\\ 
    \mathbf{K}_\mathrm{vid}=\mathbf{W}_\mathbf{K}^{(i)}\cdot \mathbf{E}_\mathrm{vid}, 
    \mathbf{V}_\mathrm{vid}=\mathbf{W}_\mathbf{V}^{(i)}\cdot \mathbf{E}_\mathrm{vid}
    \end{aligned}
    \right..
\end{equation} 
By isolating the attention pathways for video and text conditions, the model is encouraged to learn from both conditions independently, ensuring that the text condition is not overshadowed by the video condition. This design promotes a balanced and complementary integration of the two modalities, allowing the model to fully leverage the global visual semantics of the video and the localized textual semantics, as demonstrated in TExp.~\textcolor{magenta}{3}.

\vspace{-1.34em}
\subsubsection*{Observation.}
These toy experiments offer the following observations:
\begin{itemize}[leftmargin=*]
    \item[\ding{238}] Directly concatenating video condition and input latent (Struc.~\textcolor{magenta}{A}) overemphasizes video features. While this might be more effective in paired data training-based editing, it significantly limits the model's generation ability.
    \item[\ding{238}] Joint conditioning (Struc.~\textcolor{magenta}{B}) balances video and text, but the high similarity between the video condition and the denoising target often leads to video dominance, leading to insufficient editing capability.
    \item[\ding{238}] Struc.~\textcolor{magenta}{C} highlights the importance of treating video (rich visual context) and text (semantic guidance) as complementary modalities. Both must be effectively integrated for high-quality video generation and editing.
\end{itemize}

\vspace{-1.4em}
\subsubsection*{Video Condition Adaptation.} To better align text and video conditions, we leverage CLIP's~\cite{radford2021learning} visual encoder.
The tricky part, however, lies in how to encode videos with these encoders. Naive temporal aggregation (\textit{e.g.}, average pooling in TExp.~\textcolor{magenta}{2},\textcolor{magenta}{3}) loses temporal dynamics, while Video ViT \cite{bertasius2021space, wang2023internvid} is computationally costly. To address this, as illustrated in Fig.~\ref{fig:ues_cond}, we insert a lightweight adapter into the CLIP visual encoder, implemented as a depth-wise 3D convolutional layer (DWConv3D) \cite{feichtenhofer2020x3d}, inspired by \cite{pan2022st}. The adapted feature is computed as:
\setlength\abovedisplayskip{6pt}
\setlength\belowdisplayskip{6pt}
\begin{equation}
    \mathbf{X}_{ada}=\mathbf{X}+f(\text{DWConv3D}(\mathbf{X}\mathbf{W}_{down}))\mathbf{W}_{up},
\end{equation}
where $\mathbf{X}$ initially represents the patch and positional embedding of input video, $f(\cdot)$ is the activation function, and $\mathbf{W}_{down}$, $\mathbf{W}_{up}$ are the down- and up-projection layers.

Inspired by image-to-video (I2V) works \cite{xing2023dynamicrafter, chen2023videocrafter1}, we project video embeddings into a text-aligned space using the full set of video tokens $\mathbf{E}_{\mathrm{vid}}={\{\mathbf{e}^{i}\}_{i=1}^{K}}$ rather than relying solely on the global token $\mathbf{e}_{cls}$ \cite{ye2023ip, shi2024instantbooth}, to ensure comprehensive visual representation. To further align video and text embeddings, we apply a query transformer \cite{jaegle2021perceiver, jaegleperceiver} with stacked cross-attention and feed-forward layers, enabling the denoising network to effectively integrate video and text conditions. TExp.~\textcolor{magenta}{4} in Fig.~\ref{fig:toy_exp} demonstrates that our adaptation strategy preserves unedited areas significantly better than simple aggregation methods (TExp.~\textcolor{magenta}{3}).

\begin{figure}[t]
    \centering
    \vspace{-0.4em}
   \includegraphics[width=1\linewidth]{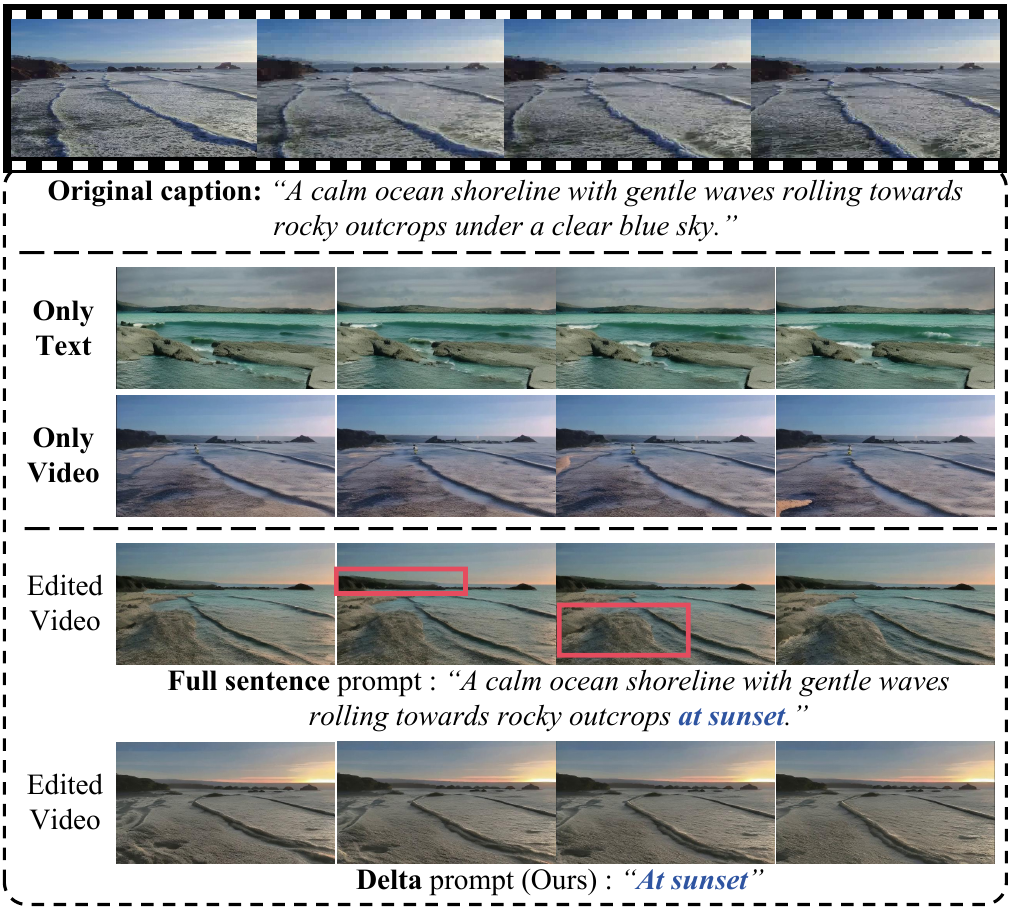}
   \vspace{-1.8em}
    \captionof{figure}{\label{fig:semantic}
    Illustration of semantic correspondence. (\textbf{\textit{Top}}) Reference video and its original caption. (\textbf{\textit{Middle}}) Results using only one condition. (\textbf{\textit{Bottom}}) Effects of full sentence \textit{vs.} delta prompt.}
    \vspace{-1.2em}
\end{figure}

\vspace{-0.2em}
\subsection{Text-Video Semantic Correspondence} 
\vspace{-0.4em}


Fig.~\ref{fig:semantic} illustrates how semantic correspondence operates in UES. The middle block highlights the limitations of relying on a single condition: \ding{182} \textit{Text-only generation} adheres to the given caption, but struggles to model global visual coherence due to the inherent ambiguities and under-specification of textual semantics. \ding{183} \textit{Video-only generation}, on the other hand, reconstructs visual details with higher fidelity, as the reference video provides comprehensive \textbf{global visual semantics}. However, it lacks the flexibility and specificity that textual guidance can offer for targeted edits.

Building upon this foundation, UES establishes a structured interaction between the global visual semantics of the reference video and the textual semantics provided by prompts, as shown in the third block of Fig.~\ref{fig:semantic}. While \textit{full-sentence prompts} encapsulate \textbf{global textual semantics}, they may conflict with the video’s global visual semantics, leading to inconsistencies or unintended alterations. In contrast, \textit{delta prompts}, which encode \textbf{local textual semantics}, focus on specific regions or temporal segments, making them more effective for guiding precise edits while preserving the video’s global structure.

This interplay between global and local semantics underscores the strength of our semantic correspondence strategy. By aligning text and video at complementary levels—global for alignment (coherence) and local for refinement (specificity)—UES achieves unified generation and editing. Moreover, this approach seamlessly adapts to diverse input contents without reliance on additional supervision signals. This ensures both flexibility and robustness across a wide range of video editing tasks.

\subsection{Fine-Tuning and Inference}
\vspace{-0.2em}
\subsubsection*{Spatiotemporal Low-Rank Adaptations.} To reduce the computational cost of UES fine-tuning, we integrate LoRAs \cite{hu2021lora} into both the spatial and temporal layers of the denoising network.
LoRAs are applied to the spatial and temporal self-attention layers as well as the feed-forward networks, while cross-attention layers remain unchanged to preserve text-video semantic correspondence.
Formally, LoRA updates the weight matrix $\mathbf{W}$ via low-rank factorization:
\begin{equation}
    \mathbf{W}=\mathbf{W}_0+\Delta \mathbf{W}=\mathbf{W}_0+BA,
\end{equation}
where $\mathbf{W}_0 \in \mathbb{R}^{d \times k}$ is the original weights, and $B \in \mathbb{R}^{d \times r}$, $A \in \mathbb{R}^{r \times k}$ are low-rank factors, with $r \ll d,k$. This enables efficient fine-tuning with UES.

\vspace{-1.3em}
\subsubsection*{Inference and Applications.} The classifier-free guidance~\cite{ho2022classifier} is naturally extended to multi-conditional during inference, \textit{i.e.}, text condition $\mathbf{c}_\mathrm{txt}$ and video condition $\mathbf{c}_\mathrm{vid}$:
\begin{equation}
    \begin{aligned}
\epsilon_\theta(\mathbf{z}_t&;t,\mathbf{c}_\mathrm{vid},\mathbf{c}_\mathrm{txt}) \leftarrow\epsilon_\theta(\mathbf{z}_t;t,\varnothing,\varnothing) \\
&+w_\mathrm{vid}\big(\epsilon_\theta(\mathbf{z}_t;t,\mathbf{c}_\mathrm{vid},\varnothing)-\epsilon_\theta(\mathbf{z}_t;t,\varnothing,\varnothing)\big) \\
&+w_\mathrm{txt}\big(\epsilon_\theta(\mathbf{z}_t;t,\mathbf{c}_\mathrm{vid},\mathbf{c}_\mathrm{txt})-\epsilon_\theta(\mathbf{z}_t;t,\mathbf{c}_\mathrm{vid},\varnothing)\big) \label{eq:cfg}
\end{aligned}
\end{equation}
where $w_\mathrm{vid}$ and $w_\mathrm{txt}$ denote the guidance scales for video and text conditions, respectively.

UES enables \textbf{unified text-based generation and editing of videos as well as images}. (i) \textit{Video}: For editing, we set $w_\mathrm{vid}$ and $w_\mathrm{txt}$ to specific values, allowing for semantic composition of the reference video and delta prompt. For generation, we use only full-sentence prompts to provide global semantics, setting $w_\mathrm{vid}=0$ to exclude the reference video. (ii) \textit{Image}: Recognizing that an image is a single frame from a video, we construct the reference video using identical reference images and sample the initial frame as the output image for both editing and generation.

\vspace{-0.6em}
\section{\texttt{OmniBench-99}: A Comprehensive Video Editing Evaluation Benchmark}
\label{sec:bench}

\vspace{-0.4em}
\subsubsection*{Overview.} Generative video editing has emerged as a rapidly growing research area, yet it still lacks a comprehensive benchmark, potentially hindering its technical advancement. Although the recently introduced BalanceCC~\cite{feng2024ccedit} organizes videos into four categories, namely, humans, animals, objects, and landscapes, it only evaluates four editing types following the LOVEU-TGVE-2023 \cite{wu2023cvpr} benchmark. This limited scope overlooks the importance of assessing editing scenarios. To this end, we present \textit{OmniBench-99}, a new benchmark of $99$ videos with text prompts that evaluate both editing types and scenarios, providing a more comprehensive benchmark for generative video editing evaluation.

\begin{figure}[t]
    \centering
    \vspace{-0.4em}
   \includegraphics[width=1\linewidth]{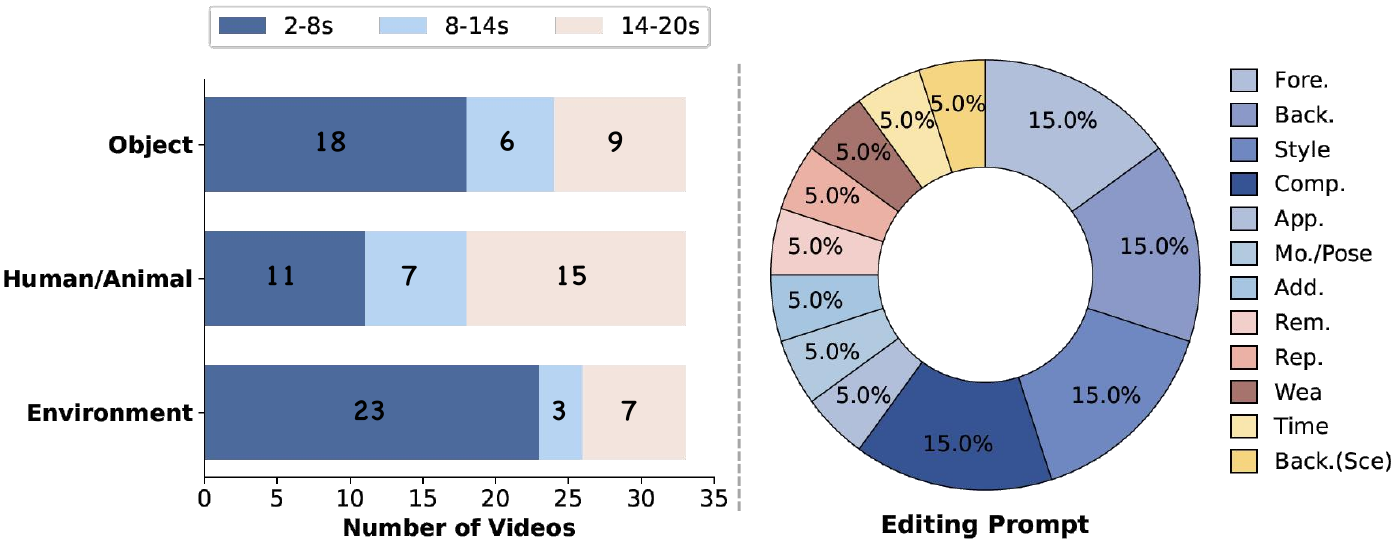}
   \vspace{-1.6em}
    \captionof{figure}{\label{fig:bench_stat}
    Statistics of OmniBench-99.}
    \vspace{-1.2em}
\end{figure}

\begin{figure*}[t]
    \centering
    \vspace{-0.4em}
   \includegraphics[width=1\linewidth]{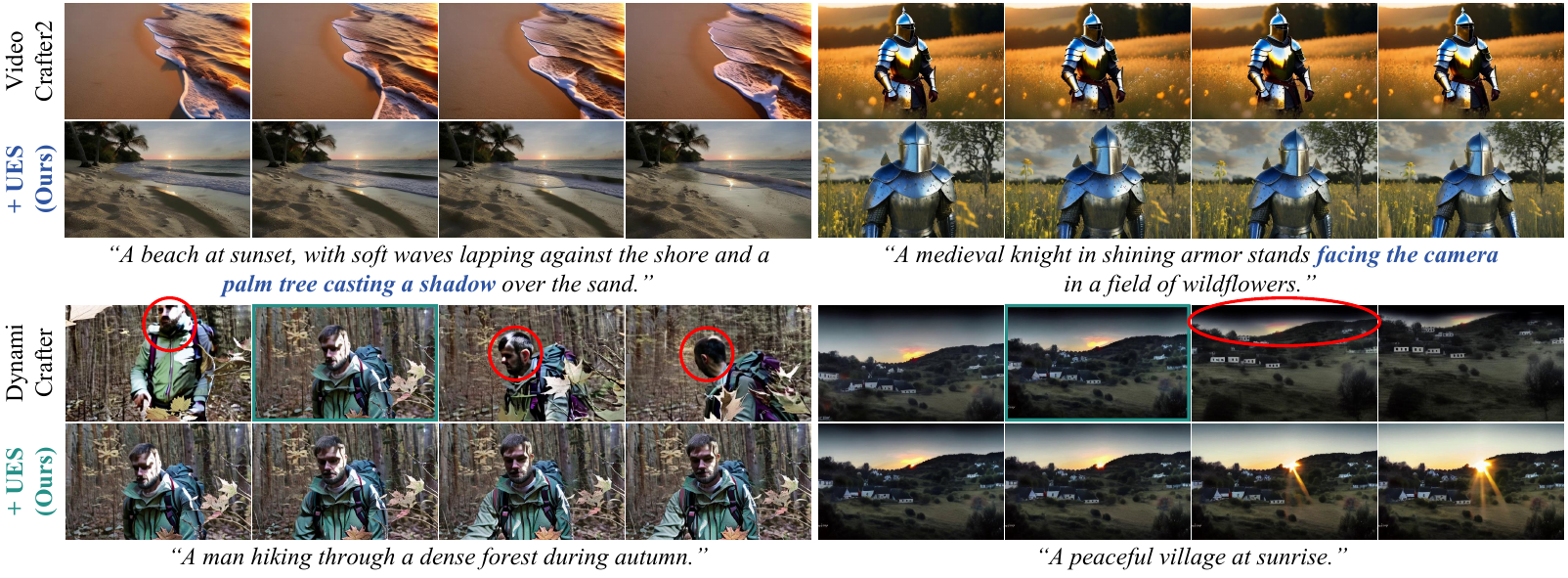}
   \vspace{-2.2em}
    \captionof{figure}{\label{fig:quali_video_gen}
    Video generation results of UES on VideoCrafter2 and DynamiCrafter. As DynamiCrafter is a T+I2V model, the first frame generated by + \textbf{\texttt{UES}} is used as the keyframe for inference. + \textbf{\texttt{UES}} enhances text alignment (Top) and temporal consistency (Bottom).}
    \vspace{-1.2em}
\end{figure*}

\vspace{-1.3em}
\subsubsection*{Establishment and Statistics.} We collected 99 open-license videos, selected for their suitability for non-stigmatizing and legal modifications. These videos range in length from 2 to 20 seconds, with a frame rate of about 30 FPS. The videos are evenly distributed across three categories: \textit{Human/Animal}, \textit{Environment}, and \textit{Object}, with 33 videos per category. In addition to generating four editing-type prompts for each video~\cite{feng2024ccedit}, we tasked GPT-4V~\cite{openai2024gpt4vsystemcard} to create category-specific prompts tailored to different editing scenarios (see Fig.~\ref{fig:example} for examples). Notably, we provide two kinds of prompts, namely \textit{full sentence} and \textit{delta caption}, for convenient use. Afterward, we conducted a thorough manual inspection to ensure the quality of these prompts. We present the statistical distribution of OmniBench-99 in Fig.~\ref{fig:bench_stat}. We hope this new benchmark will better address gaps in previous research and provide a reliable standard for generative video editing. More details and visual examples are available in \textbf{Appendix}~\S\ref{app_bench}.

\begin{figure}[t]
    \centering
    \vspace{-0.2em}
   \includegraphics[width=0.98\linewidth]{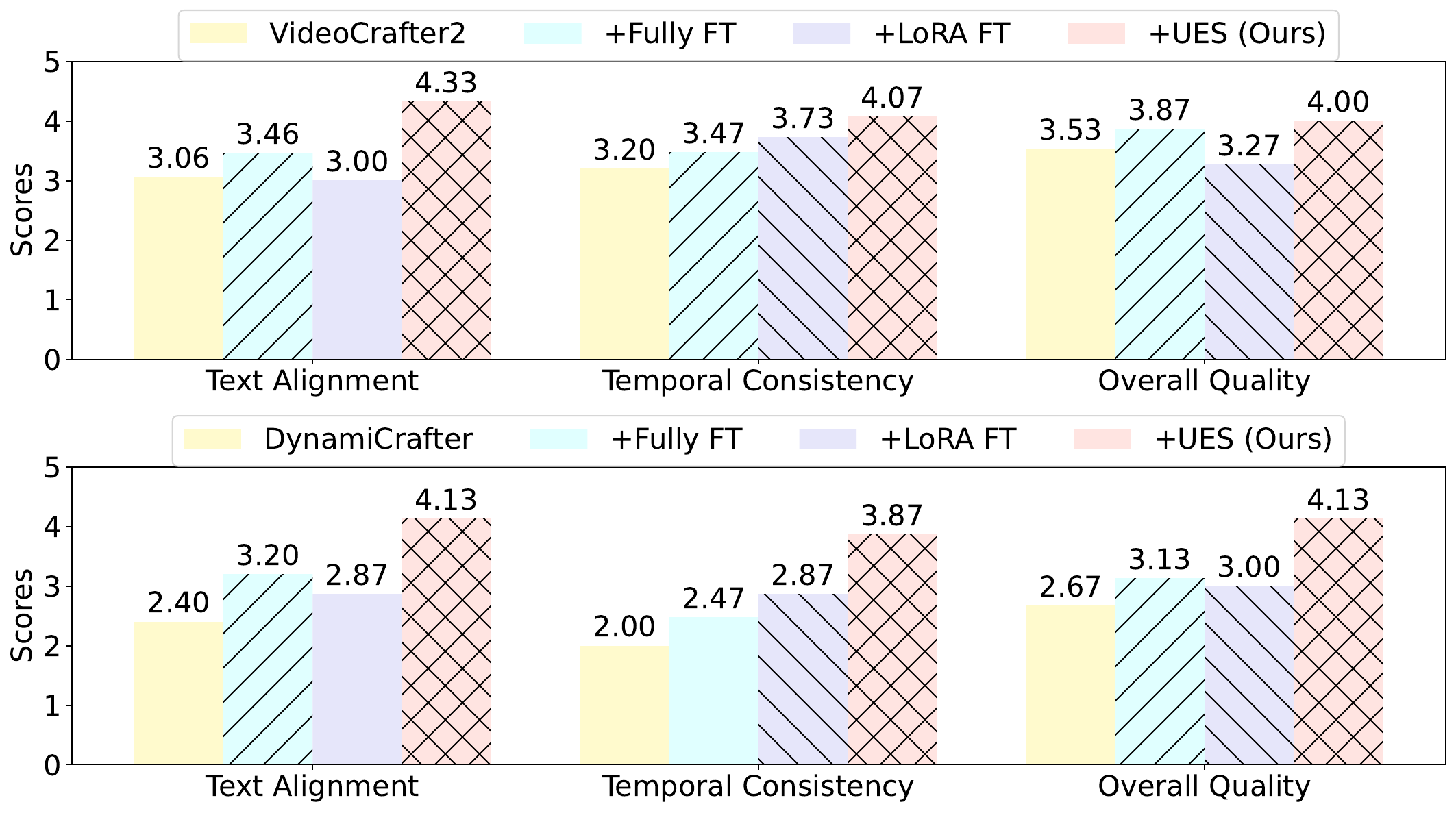}
   \vspace{-0.8em}
    \captionof{figure}{\label{fig:user_study}
    User studies on T(+I)2V generation of UES.}
    \vspace{-0.2em}
\end{figure}

\begingroup
\setlength{\tabcolsep}{2.4pt}
\begin{table}[t]
\renewcommand{\arraystretch}{1}
  \centering
  \caption{Comparison with T(+I)2V generation models on VBench. Results across all $16$ dimensions are provided in \textbf{Appendix}~\S\ref{app_more_gen}.}
  \centering
  \vspace{-0.8em}
  \scriptsize
   \begin{tabular}{lcccccccc}
     \hlineB{2.5}
     \textbf{Method} & \textbf{Total} & \textbf{QS} & \textbf{SS} & \textbf{HA} & \textbf{Scene} & \textbf{DD} & \textbf{MO} & \textbf{AS}  \\
     \hlineB{1.5}
     VideoCrafter2~\cite{chen2024videocrafter2} & 80.44 & 82.20 & 73.42 & \textbf{95.00} & 55.29 & \underline{42.50} & 40.66 & \textbf{25.13} \\
     $+$ Fully Fine-tune & \underline{80.81} & \underline{82.57} & \underline{73.78} & \underline{94.34} & 54.12 & \textbf{42.58} & 42.11 & \underline{25.02}\\
     $+$ LoRA Fine-tune & 80.49 & 82.29 & 73.30 & 93.12 & \underline{55.87} & 42.32 & \underline{44.34} & 24.87\\
     \rowcolor{cyan!10}
     $+$ \textbf{\texttt{UES} (Ours)} & \textbf{81.90} & \textbf{83.51} & \textbf{75.45} & 93.20 & \textbf{56.12} & 42.40 &  \textbf{53.43} & 24.98 \\
     \hline
     DynamiCrafter~\cite{xing2023dynamicrafter} & 80.60 & 82.74 & 72.09 & \textbf{94.56} & 56.24 & \underline{69.67} & 38.98 & \underline{23.12}\\
     $+$ Fully Fine-tune & 80.42 & 82.52 & 72.03 & \underline{94.40} & 56.50 & 69.50 & 39.50 & \textbf{23.30} \\
     $+$ LoRA Fine-tune & \underline{80.78} & \underline{82.89} & \underline{72.32} & 94.00 & \underline{56.80} & 69.40 & \underline{40.20} & 23.00 \\
     \rowcolor{cyan!10}
     $+$ \textbf{\texttt{UES} (Ours)} & \textbf{81.73} & \textbf{83.71} & \textbf{73.76} & 93.80 &  \textbf{57.20} & \textbf{69.80} & \textbf{44.50} & \underline{23.12} \\
     \hlineB{2.5}
   \end{tabular}
\parbox{0.48\textwidth}{
{\scriptsize \textbf{Note:} QS: Quality Score, SS: Semantic Score, HA: Human Action, DD: Dynamic Degree, MO: Multiple Objects, AS: Appearance Style.}
}
  \label{tab:quantitative_vbench}
  \vspace{-0.3cm}
\end{table} 
\endgroup

\section{Experiment}
\label{sec:experi}

\vspace{-0.4em}
In this section, we empirically investigate the effectiveness of UES in unlocking universal editing capability while empowering the generation capability.

\subsection{Experimental Settings}
\vspace{-0.2em}
\subsubsection*{Baselines.} We apply UES to two distinct video generators: T+I2V model DynamiCrafter-512 \cite{xing2023dynamicrafter} and T2V model VideoCrafter2-512 \cite{chen2024videocrafter2}. \textit{To highlight UES’s superiority, we fine-tune the base models following their default settings (e.g., resolution, frame length). Notably, UES is compatible with most text(+image)-to-video model.} 
For video editing, we compare + \texttt{\textbf{UES}} with CCEdit \cite{feng2024ccedit}, Video-P2P \cite{liu2024video}, InsV2V \cite{cheng2023consistent}, TokenFlow \cite{geyer2023tokenflow}, ControlVideo \cite{zhao2023controlvideo}, and Tune-A-Video \cite{wu2023tune}. For generation, while other vanilla generation models cannot be directly compared to models enhanced with UES, we compare + \texttt{\textbf{UES}} with fully fine-tuning and LoRA fine-tuning. 

\begin{figure}[t]
    \centering
   \includegraphics[width=1\linewidth]{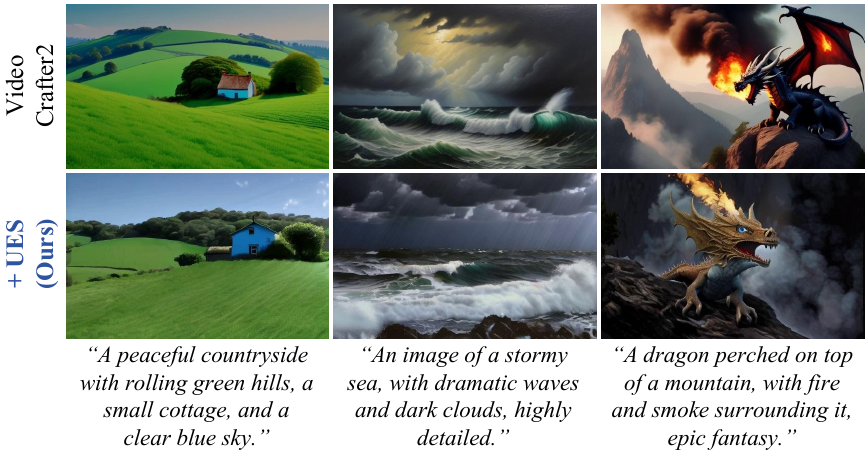}
   \vspace{-1.8em}
    \captionof{figure}{\label{fig:quali_t2i}
    Image generation results of UES on VideoCrafter2. +~\textbf{\texttt{UES}} generates more realistic images than base model.}
    \vspace{-1.2em}
\end{figure}

\begin{figure*}[t]
    \centering
    \vspace{-0.4em}
   \includegraphics[width=1\linewidth]{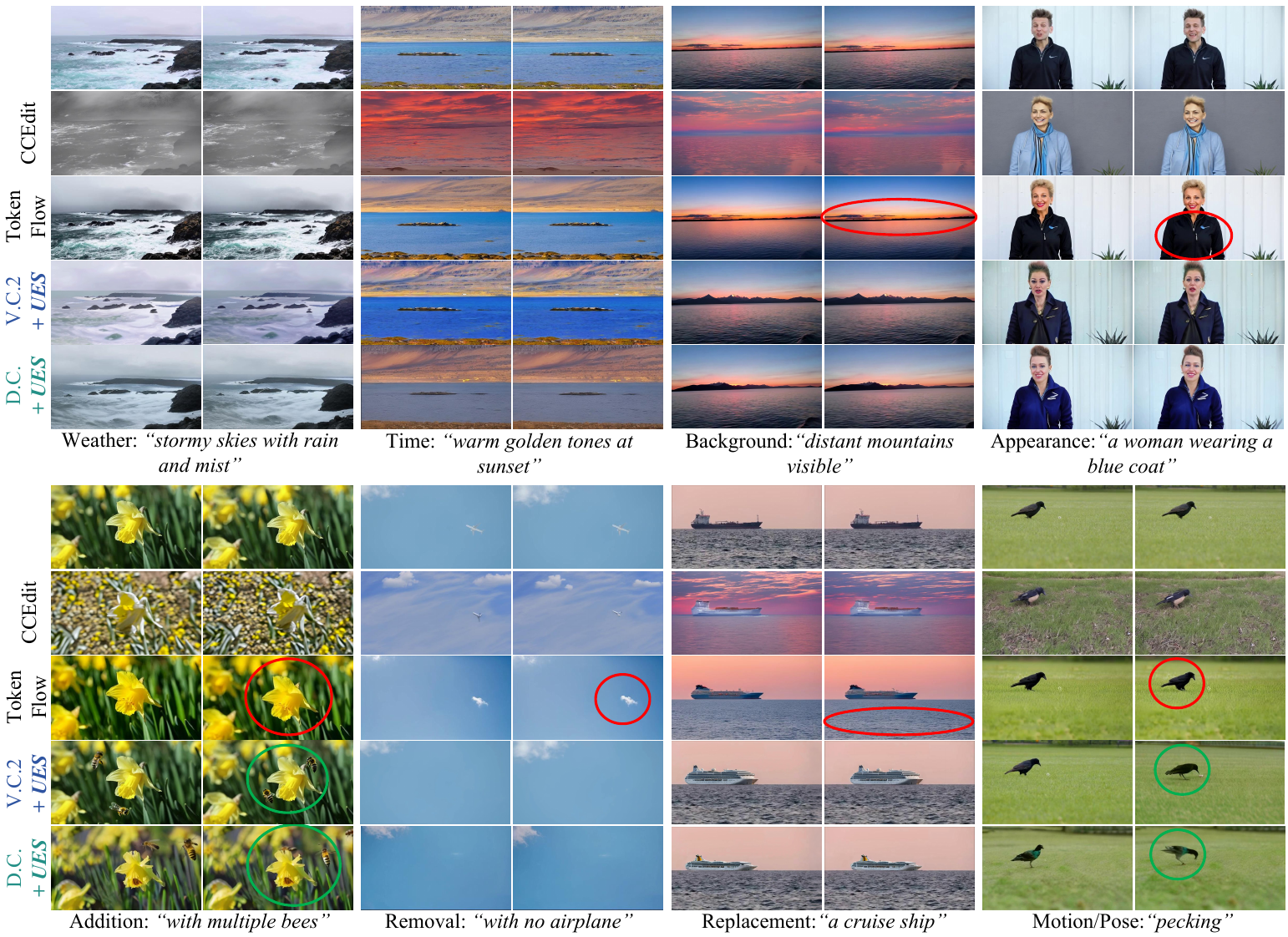}
   \vspace{-2em}
    \captionof{figure}{\label{fig:quali_video_edit}
    Video editing comparison of UES on VideoCrafter2 (V.C.2) and DynamiCrafter (D.C.). We follow the baseline's prompt setting but only show delta prompts here. For complete comparisons, please refer to \textbf{Appendix}~\S\ref{app_more_edit}.}
    \vspace{-0.4em}
\end{figure*}

\begingroup
\setlength{\tabcolsep}{2.4pt}
\begin{table*}[t]
\renewcommand{\arraystretch}{1}
  \centering
  \caption{Quantitative comparison with text-guided video editing methods on our OmniBench-99. We also conduct a quantitative evaluation on the BalanceCC~\cite{feng2024ccedit} and LOVEU-TGVE-2023 \cite{wu2023cvpr} benchmarks in \textbf{Appendix}~\S\ref{app_more_edit}.}
  \centering
  \vspace{-0.8em}
  \scriptsize
   \begin{tabular}{lc|cc|cccc|cc|cccc}
     \hlineB{2.5}
     \multirow{3}{*}{\textbf{Method}} & \textbf{Additional} & \multicolumn{6}{c|}{\textbf{Editing Type}} & \multicolumn{6}{c}{\textbf{Eidting Scenario}}\\
     \cline{3-8}\cline{9-14}
      & \textbf{Supervision} & \multicolumn{2}{c|}{\textbf{Automatic}} & \multicolumn{4}{c|}{\textbf{User Study}} & \multicolumn{2}{c|}{\textbf{Automatic}} & \multicolumn{4}{c}{\textbf{User Study}}\\
     \cline{3-8}\cline{9-14}
     & \textbf{Need} & \textbf{CLIP Frame$\uparrow$} & \textbf{PickScore$\uparrow$} & \textbf{Align.$\uparrow$} & \textbf{Temp.$\uparrow$} & \textbf{Stru.$\uparrow$} & \textbf{Overall$\uparrow$} & \textbf{CLIP Frame $\uparrow$} & \textbf{PickScore$\uparrow$} &  \textbf{Align.$\uparrow$} & \textbf{Temp.$\uparrow$} & \textbf{Stru.$\uparrow$} & \textbf{Overall$\uparrow$} \\
     \hlineB{2}
     Tune-A-Video~\cite{wu2023tune} & \textcolor{red}{\ding{51}} & 0.931 & 0.205 & 3.07  & 2.87 & 3.13 & 3.20 & 0.929 & 0.198 & 3.33  & 2.93 & 3.00 & 3.13\\
     ControlVideo~\cite{zhao2023controlvideo} & \textcolor{red}{\ding{51}}   & 0.949 & 0.210 & 2.93  & 2.27 & 2.40 & 2.40 & 0.950 & 0.203 &  1.80 & 1.87 & 2.13 & 2.27\\
     TokenFlow~\cite{geyer2023tokenflow} & \textcolor{red}{\ding{51}}   & 0.948 & 0.208 & 2.73  & 3.33 & 2.80 & 3.07 & 0.951 & 0.200 &  3.07 & 3.07 & 2.93 & 3.13\\
     InsV2V~\cite{cheng2023consistent} & \textcolor{red}{\ding{51}}   & 0.914 & 0.208 & 2.13  & 2.20 & 2.33 & 2.47 & 0.911 & 0.198 &  1.73 & 1.93 & 1.87 & 2.00\\
     Video-P2P~\cite{liu2024video} & \textcolor{red}{\ding{51}}   & 0.930 & 0.198 &  3.13 & 3.27 & 3.20 & 3.00 & 0.928 & 0.189 & 3.13  & 3.20 & 3.13 & 3.07\\
     CCEdit~\cite{feng2024ccedit} & \textcolor{red}{\ding{51}}   & 0.932 & 0.210 &  1.73 & 2.53 & 2.27 & 2.20 & 0.935 & 0.204 &  1.53 & 2.53 & 2.20 & 2.20\\
     \hline
     \rowcolor{lightgray!20}
     VideoCrafter2~\cite{chen2024videocrafter2} & \multicolumn{13}{c}{{\textit{No Editing Capability}}}\\
     \rowcolor{cyan!10}
     $+$ \textbf{\texttt{UES} (Ours)} & \textcolor{green}{\ding{55}}  &  \textbf{0.966} & \underline{0.211} & \underline{4.33} & \textbf{4.40} & \underline{4.13} & \underline{4.00} & \textbf{0.969} & \underline{0.209} & \underline{4.00}  & \textbf{4.27} & \textbf{4.20} & \underline{3.93}\\
     \hline
     \rowcolor{lightgray!20}
     DynamiCrafter~\cite{xing2023dynamicrafter} & \multicolumn{13}{c}{{\textit{No Editing Capability}}}\\
     \rowcolor{cyan!10}
     $+$ \textbf{\texttt{UES} (Ours)} & \textcolor{green}{\ding{55}}  &  \underline{0.962} & \textbf{0.212} & \textbf{4.47} & \underline{4.33} & \textbf{4.07} & \textbf{4.33} & \underline{0.966} & \textbf{0.216} & \textbf{4.07}  & \underline{4.13} & \textbf{4.20} & \textbf{4.00}\\
     \hlineB{2.5}
   \end{tabular}
  \label{tab:quantitative}
  \vspace{-1em}
\end{table*} 
\endgroup

\vspace{-1.3em}
\subsubsection*{Evaluations.}
(i) \textbf{Automatic Metrics}: For video editing, we use PickScore \cite{kirstain2023pick} to evaluate the alignment between all frames of the output video and the corresponding edited prompt. Additionally, we employ CLIP Frame Consistency \cite{radford2021learning} to measure the average cosine similarity between CLIP image embeddings across all frames. We adapt VBench \cite{huang2023vbench} for generation.
(ii) \textbf{User Study}: We conduct a user study using the mean opinion score (MOS) metric, focusing on four key aspects: Text Alignment, Temporal Consistency, Structure Alignment (specific to editing), and Overall Quality.
(iii) \textbf{Benchmarks}: For video editing, we evaluate using our newly constructed OmniBench-99 benchmark, which comprehensively covers both editing types and scenarios. For generation, we assess on VBench benchmark.

\vspace{-1.5em}
\subsubsection*{Implementation Details.}
We adopt a high-quality T2V dataset, OpenVidHD-0.4M \cite{nan2024openvid} for UES fine-tuning, using $4$ NVIDIA H20 GPUs for $12$K steps with a batch size of $32$. Each video is sampled with $16$ frames at a resolution of $320\times512$, following the \textbf{default settings} of the base models.

\subsection{Enhanced Generation}

\vspace{-0.4em}
Before demonstrating how UES unlocks universal editing, we first show its enhancement to the model's original generation ability, ensuring editing does not degrade generation.

\vspace{-1.4em}
\subsubsection*{Quantitative Comparison.} We evaluate UES on VBench \cite{huang2023vbench} comparing it with fully fine-tuning and LoRA fine-tuning under the same steps to provide a clear comparison. As shown in Tab.~\ref{tab:quantitative_vbench}, UES achieves the best performance improvement with minimal tunable parameters. User studies (Fig.~\ref{fig:user_study}) further confirm that UES improves text alignment, temporal consistency, and overall quality.

\vspace{-1.6em}
\subsubsection*{Qualitative Comparison.} To illustrate the enhancement brought by UES more vividly, we present qualitative video comparisons in Fig.~\ref{fig:quali_video_gen} on VideoCrafter2 and DynamicCrafter. It is evident that models fine-tuned with UES exhibit more superior temporal consistency and text alignment.
\textbf{Note:} Since base settings (\textit{e.g.}, resolution, frame length) are preserved for a fair comparison, video quality and motion dynamics largely depend on the base model itself. These aspects can be further improved with stronger base models.
Additionally, we compare image generation results in Fig.~\ref{fig:quali_t2i}, which show that UES enhances realism. 

\begin{figure}[t]
    \centering
    \vspace{-0.6em}
   \includegraphics[width=1\linewidth]{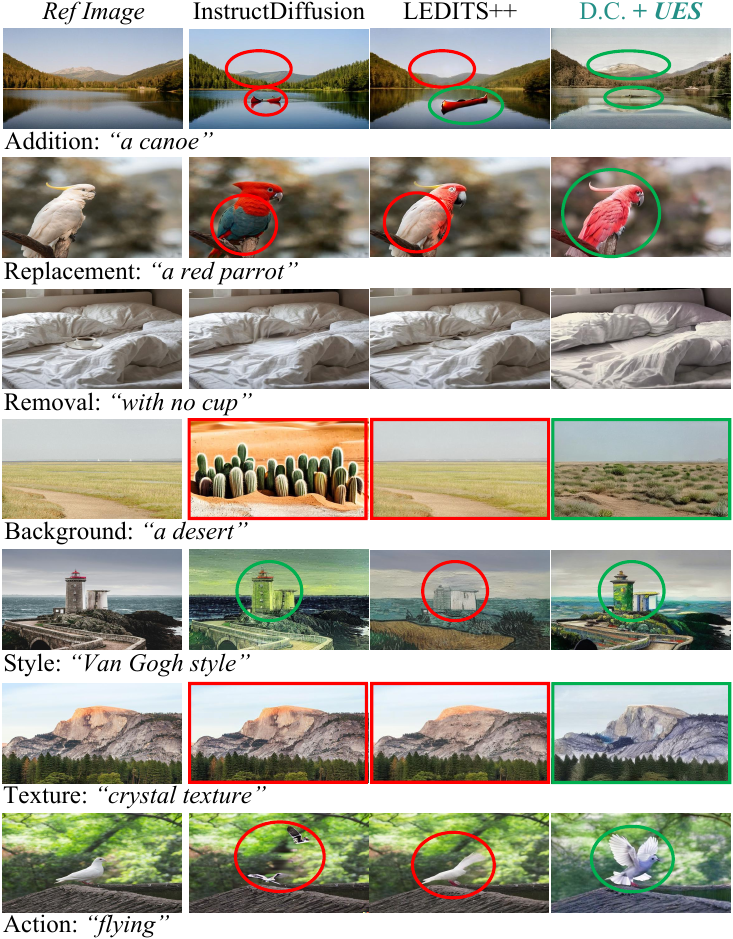}
   \vspace{-1.8em}
    \captionof{figure}{\label{fig:quali_image_edit}
    Image editing comparison of UES on DynamiCrafter on on EditEval \cite{huang2024diffusion}. More results are provided in \textbf{Appendix}~\S\ref{app_more_imgedit}.}
    \vspace{-1.2em}
\end{figure}

\vspace{-0.1em}
\subsection{Universal Editing}
\vspace{-0.1em}
UES unlocks the universal editing capabilities of video generators. To thoroughly evaluate this, we adopt our proposed \textbf{\texttt{OmniBench-99}}, which extends previous benchmarks \cite{feng2024ccedit, wu2023cvpr} (solely focused on $4$ editing types: Foreground, Background, Composite, Overall/Style) by covering $8$ editing scenarios: (\textit{i}) \textbf{Environment}: Weather, Time, Background. (\textit{ii}) \textbf{Object}: Addition, Removal, Replacement. (\textit{iii}) \textbf{Human/Animal}: Appearance, Motion/Pose.

\vspace{-1.2em}
\subsubsection*{Quantitative Results.} As shown in Tab.~\ref{tab:quantitative}, while SotA methods achieve competitive performance by relying on additional supervision, it is noteworthy that UES enables models without inherent editing capabilities to perform powerful and universal editing without any extra supervision. User studies further confirm its effectiveness.

\vspace{-1.4em}
\subsubsection*{Qualitative Results.} Due to space limitations, we present comparisons for $8$ editing scenarios in Fig.~\ref{fig:quali_video_edit}, using two baselines (CCEdit \cite{feng2024ccedit} and TokenFlow \cite{geyer2023tokenflow}). 
While baselines perform well in specific cases, their reliance on supervision limits generalization. In contrast, models with UES exhibit strong editing performance across all $8$ scenarios, underscoring its superiority.
We also showcase UES's image editing comparisons against SotA models (InstructDiffusion \cite{geng2024instructdiffusion}, LEDITS++ \cite{brack2024ledits++}). Results on a comprehensive image editing benchmark EditEval \cite{huang2024diffusion} further validate that \textbf{\textit{UES unifies editing, generation, and video and image tasks}}. Qualitative results are provided in \textbf{Appendix}~\S\ref{app_more_imgedit}.

\begin{figure}[t]
    \centering
    \vspace{-0.2em}
   \includegraphics[width=1\linewidth]{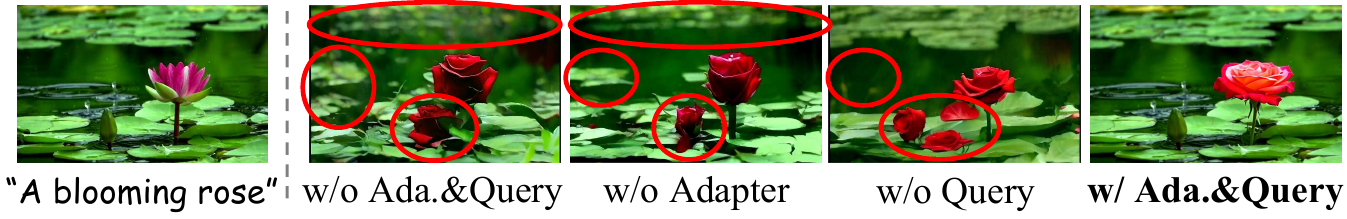}
   \vspace{-1.8em}
    \captionof{figure}{\label{fig:abla_compo}
    Ablation study on UES condition modeling. Including the adapter and query transformer significantly contributes to achieving precise editing capabilities.}
    \vspace{-0.6em}
\end{figure}

\begin{figure}[t]
    \centering
   \includegraphics[width=1\linewidth]{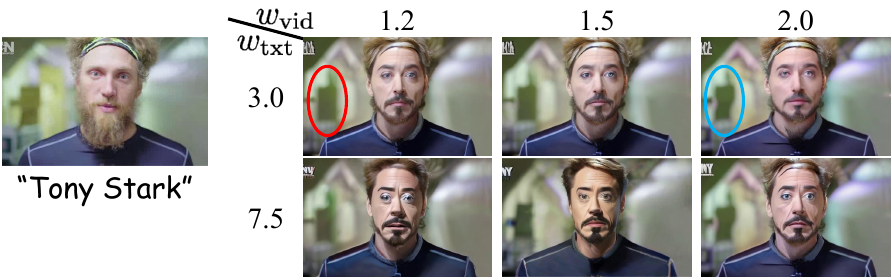}
   \vspace{-1.8em}
    \captionof{figure}{\label{fig:abla_cfg}
    Ablation study on multimodal guidance. $w_\mathrm{txt}$ controls consistency with the edit instruction, while $w_\mathrm{vid}$ controls the similarity with reference video.}
    \vspace{-1.2em}
\end{figure}

\vspace{-0.2em}
\subsection{Ablation Studies}

\vspace{-0.2em}
\subsubsection*{Condition Modeling.} We evaluate UES's condition modeling: the adapter and query transformer in Fig.~\ref{fig:abla_compo}. Row 3 \textit{vs.} row 5 shows the adapter ensures overall consistency, while row 4 \textit{vs.} row 5 shows the query transformer refines local features for better detail accuracy and fidelity.

\vspace{-1.4em}
\subsubsection*{Multimodal Guidance.} As shown in Fig.~\ref{fig:abla_cfg}, increasing multimodal CFG scales enhances control over editing results. Notably, higher video condition scale $w_\mathrm{vid}$ not only improves control over facial regions but also better preserves the spatial structure of the reference's background.
\vspace{-1.4em}

\section{Conclusion}
\label{sec:conclu}
\vspace{-0.4em}
In this work, we propose \textbf{\texttt{UES}} (Unlocking Universal Editing via Self-Supervision), a lightweight self-supervised fine-tuning strategy that transforms video generation models into unified generation-editing systems through self-supervised semantic alignment. Leveraging a dual-conditioning mechanism with video-text pairs, UES enables structured learning of intrinsic spatiotemporal correspondences. It offers three key advantages: Universality, Unification, and Efficiency. To validate UES, we introduce \textbf{\texttt{OmniBench-99}}, a benchmark spanning 99 videos across four editing types and eight scenarios. Extensive experiments show that UES equips models without inherent editing capabilities to perform versatile and precise editing while preserving or enhancing their generation performance.
{
    \small
    \bibliographystyle{ieeenat_fullname}
    \bibliography{main}
}
\clearpage
\maketitlesupplementary
\appendix
\setcounter{tocdepth}{0}
\tableofcontents
\addtocontents{toc}{\setcounter{tocdepth}{2}}

\vspace{4mm}
\noindent \textit{The code will be fully released along with more base model implementations in June.}

\section{Editing Capabilities Overview}
\label{app_overview}

\begin{figure}[t]
    \centering
   \includegraphics[width=1\linewidth]{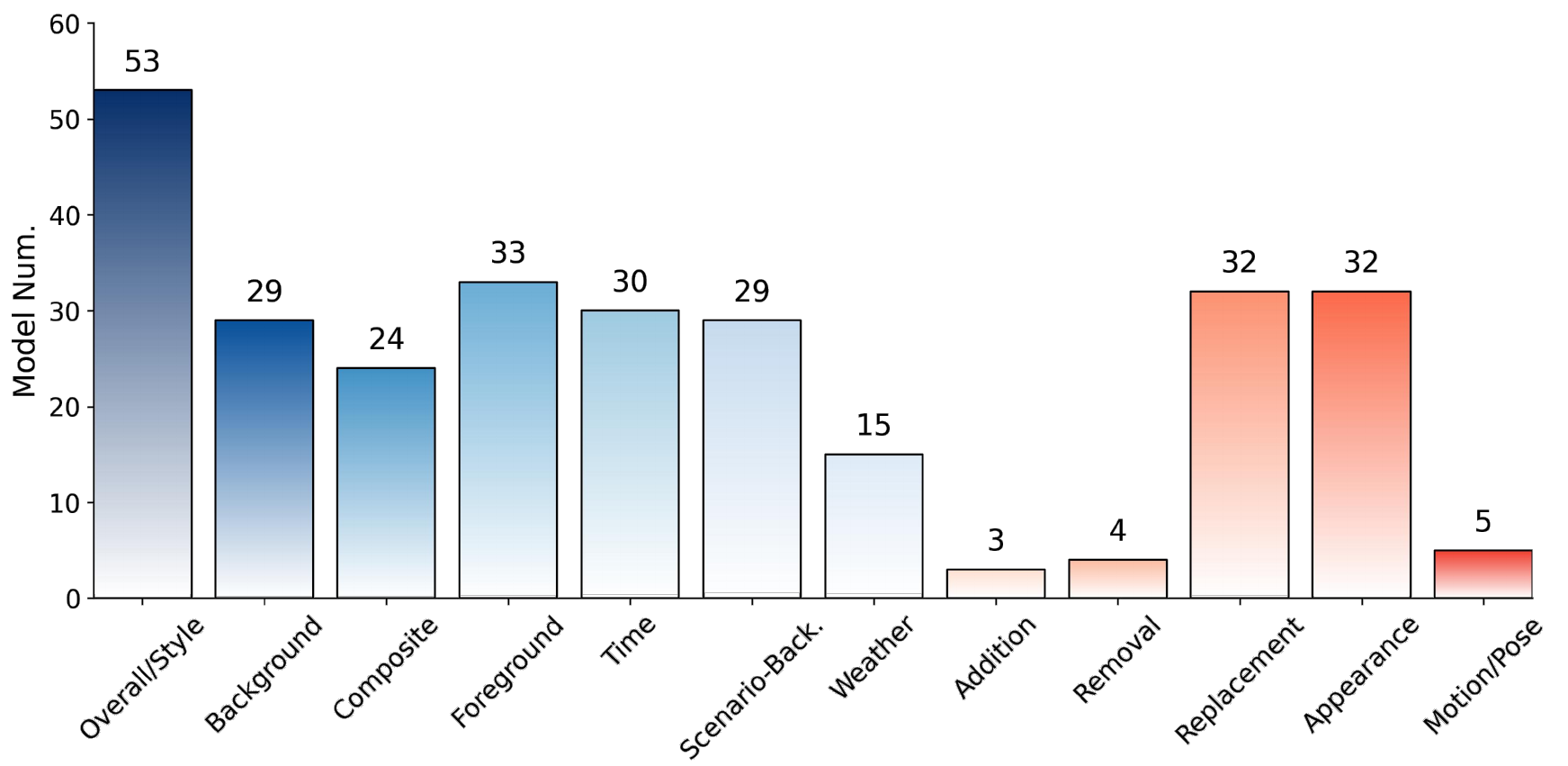}
   \vspace{-1.8em}
    \captionof{figure}{\label{sup_fig:edit_stat}
    Editing capabilities statistical overview of $57$ models: number of models corresponding to each editing. Despite relying on additional supervision signals, existing editing models exhibit varying levels of performance and generalization, often with notable limitations. In contrast, \textbf{\texttt{UES}} leverages self-supervised fine-tuning to unlock the universal editing capabilities of generative models, offering a more robust and versatile solution. More details in Table~\ref{app_tab:review}.}
    \vspace{-0.4em}
\end{figure}

To facilitate a clearer understanding of the current advancements in text-based video editing, we perform a comprehensive evaluation and review, as illustrated in Figure~\ref{sup_fig:edit_stat} and Table~\ref{app_tab:review}, structured around the following aspects:

\vspace{-0.8em}
\subsubsection*{Tune \& I2I.} The ``Tune" refers to techniques like one-video tuning, as seen in Tune-A-Video~\cite{wu2023tune}, or few-video tuning in MotionDirector~\cite{zhao2023motiondirector}. While this method helps models learn the specific structure of given videos, it tends to limit the model's generalizability and flexibility in practical applications; The ``I2I" method involves using image-to-image models, \textit{e.g.}, IP2P~\cite{brooks2023instructpix2pix}, to edit individual video frames. Although this allows for precise edits at the image level, it introduces challenges like increased computational cost, potential inconsistencies, and uncertainty in the editing process.

\begingroup
\setlength{\tabcolsep}{2.4pt}
\begin{table*}[t]
\vspace{0.6em}
\renewcommand{\arraystretch}{1}
  \centering
  \caption{\textbf{Editing Capability Overview.} \textbf{``Tune":} One-shot or few-shot tuning-based; \textbf{``I2I":} Image editing model (\textit{e.g.}, InstructPix2Pix~\cite{brooks2023instructpix2pix}) assisted; \textbf{``Additional Control":} `Cond.': Condition=\{Ma=Mask, Mo=Motion vector, E=Edge, O=Optical flow, C=Canny, H=HED boundary, D=Depth, P=Pose, S=Sketch, SI=Style Image, B=Bounding box, A=Atlas\}, `Feat.': Attention feature injection during inference, `DDIM': DDIM inversion-assisted;
  \textbf{``Editing Type":} `Fore.': Foreground, `Back.': Background, `Comp.': Composite, `Overall': only for overall editing, \textit{e.g.}, style;
  \textbf{``Editing Scenario":} `App.': Appearance, `Mo.': Motion, `Add.': Addition, `Rem.': Removal, `Rep.': Replacement, `Wea.': Weather, `Back.': Background. \textit{Note: since many methods are not open source, we only evaluate this type of model through the results shown in its paper/page.}}
  \centering
  \vspace{-0.8em}
  \scriptsize
   \begin{tabular}{|l|cc|ccc|cccc|cc|ccc|ccc|}
     \hlineB{2.5}
     \multirow{3}{*}{\textbf{Method}} & \multirow{3}{*}{\textbf{Tune}} & \multirow{3}{*}{\textbf{I2I}} & \multicolumn{3}{c|}{\multirow{2}{*}{\textbf{Additional Control}}} & \multicolumn{4}{c|}{\multirow{2}{*}{\textbf{Editing Type}}} & \multicolumn{8}{c|}{\textbf{Editing Scenario}}\\
     \cline{11-18}
     & & &  & & &  &  &  & & \multicolumn{2}{c|}{\textbf{Human/Animal}} &  \multicolumn{3}{c|}{\textbf{Object}} &  \multicolumn{3}{c|}{\textbf{Environment}}\\
     \cline{4-6}\cline{7-10}\cline{11-18}
     & & & \textbf{Cond.} & \textbf{Feat.} & \textbf{DDIM} & \textbf{Fore.} & \textbf{Back.} & \textbf{Comp.} & \textbf{Overall} & \textbf{App.} & \textbf{Mo./Pose} & \textbf{Add.} & \textbf{Rem.} & \textbf{Rep.} & \textbf{Wea.} & \textbf{Time} & \multicolumn{1}{c|}{\textbf{Back.}}\\
     \hlineB{2}
     \rowcolor{yellow!20} \multicolumn{18}{|c|}{{\textit{\textbf{Video-Video Pair-based}}}} \\
     InstuctVid2Vid~\cite{qin2023instructvid2vid} \textit{\textcolor{gray}{{\fontsize{6}{12}\selectfont ICME'24}}}& &  &  &  &  & \ding{51} & \ding{51} &  & \ding{51} & \ding{51} & &  &  & \ding{51} &  \ding{51} & \ding{51} & \ding{51}\\
     InsV2V~\cite{cheng2023consistent} \textit{\textcolor{gray}{{\fontsize{6}{12}\selectfont ICLR'24}}} & &  &  &  &  & \ding{51} & \ding{51} & \ding{51} & \ding{51} & \ding{51} & &  &   & \ding{51} & \ding{51} & \ding{51} & \ding{51}\\
     \hline
     \rowcolor{yellow!20} \multicolumn{18}{|c|}{{\textit{\textbf{Text-Video Pair-based}}}} \\
     \rowcolor{gray!20} \multicolumn{18}{|c|}{{\textit{Only Style/Overall Editing}}} \\
     Control-A-Video~\cite{chen2023control} \textit{\textcolor{gray}{{\fontsize{6}{12}\selectfont arXiv'23}}} &  &  & C/H/D  & & &   &  & & \ding{51} & \cellcolor{gray!20} & \cellcolor{gray!20} & \cellcolor{gray!20} & \cellcolor{gray!20} & \cellcolor{gray!20} & \cellcolor{gray!20} &  \cellcolor{gray!20} & \cellcolor{gray!20} \\
     Video ControlNet~\cite{chu2023video} \textit{\textcolor{gray}{{\fontsize{6}{12}\selectfont arXiv'23}}} & &   & O/D &   & &  & & & \ding{51}  & \cellcolor{gray!20} & \cellcolor{gray!20} & \cellcolor{gray!20} & \cellcolor{gray!20} & \cellcolor{gray!20} & \cellcolor{gray!20} & \cellcolor{gray!20} & \cellcolor{gray!20} \\
     VideoControlNet~\cite{hu2023videocontrolnet} \textit{\textcolor{gray}{{\fontsize{6}{12}\selectfont arXiv'23}}} &  &   &O+Ma/D/C &  & &  &  & & \ding{51} & \cellcolor{gray!20} & \cellcolor{gray!20} & \cellcolor{gray!20} & \cellcolor{gray!20} &  \cellcolor{gray!20} & \cellcolor{gray!20} & \cellcolor{gray!20} & \cellcolor{gray!20} \\
     Dreamix~\cite{molad2023dreamix} \textit{\textcolor{gray}{{\fontsize{6}{12}\selectfont arXiv'23}}} & \ding{51} &  & &  & \ding{51} &   & & & \ding{51} & \cellcolor{gray!20} & \cellcolor{gray!20} & \cellcolor{gray!20} & \cellcolor{gray!20} & \cellcolor{gray!20} &  \cellcolor{gray!20} & \cellcolor{gray!20} & \cellcolor{gray!20} \\
     Vid2Vid-Zero~\cite{wang2023zero} \textit{\textcolor{gray}{{\fontsize{6}{12}\selectfont arXiv'23}}} &  &  & & \ding{51} & \ding{51} &  & &  & \ding{51} &\cellcolor{gray!20} & \cellcolor{gray!20} & \cellcolor{gray!20} & \cellcolor{gray!20} &  \cellcolor{gray!20} & \cellcolor{gray!20} & \cellcolor{gray!20} & \cellcolor{gray!20} \\
     Fate-Zero~\cite{qi2023fatezero} \textit{\textcolor{gray}{{\fontsize{6}{12}\selectfont ICCV'23}}}& &  & & \ding{51}  & \ding{51} &   & & & \ding{51} & \cellcolor{gray!20} & \cellcolor{gray!20} & \cellcolor{gray!20} &  \cellcolor{gray!20} & \cellcolor{gray!20} & \cellcolor{gray!20} & \cellcolor{gray!20} & \cellcolor{gray!20} \\
     Pix2Video~\cite{ceylan2023pix2video} \textit{\textcolor{gray}{{\fontsize{6}{12}\selectfont ICCV'23}}} &  & \ding{51} &  & & \ding{51} &  & &  & \ding{51} & \cellcolor{gray!20} & \cellcolor{gray!20} & \cellcolor{gray!20} & \cellcolor{gray!20} & \cellcolor{gray!20} & \cellcolor{gray!20} & \cellcolor{gray!20} & \cellcolor{gray!20} \\
     EI$^2$~\cite{zhang2024towards} \textit{\textcolor{gray}{{\fontsize{6}{12}\selectfont NeurIPS'23}}}& \ding{51}  & &  &  & \ding{51}  &  & & & \ding{51} & \cellcolor{gray!20} & \cellcolor{gray!20} & \cellcolor{gray!20} &  \cellcolor{gray!20} & \cellcolor{gray!20} & \cellcolor{gray!20} & \cellcolor{gray!20} & \cellcolor{gray!20} \\
     RAV~\cite{yang2023rerender} \textit{\textcolor{gray}{{\fontsize{6}{12}\selectfont SIGGRAPH Asia'23}}} &  & \ding{51} &  E+O &  &  &  & & & \ding{51} & \cellcolor{gray!20} & \cellcolor{gray!20} &  \cellcolor{gray!20} & \cellcolor{gray!20} & \cellcolor{gray!20} & \cellcolor{gray!20} & \cellcolor{gray!20} & \cellcolor{gray!20}  \\
     MotionClone~\cite{ling2024motionclone} \textit{\textcolor{gray}{{\fontsize{6}{12}\selectfont arXiv'24}}} & &  &  &  & \ding{51} &  &  &  & \ding{51} & \cellcolor{gray!20} & \cellcolor{gray!20} & \cellcolor{gray!20} & \cellcolor{gray!20} & \cellcolor{gray!20} & \cellcolor{gray!20} & \cellcolor{gray!20} & \cellcolor{gray!20} \\
     Make-Your-Video~\cite{xing2024make} \textit{\textcolor{gray}{{\fontsize{6}{12}\selectfont TVCG'24}}} & &  & D  &  & &   &  &  & \ding{51} & \cellcolor{gray!20} & \cellcolor{gray!20} & \cellcolor{gray!20} &  \cellcolor{gray!20} & \cellcolor{gray!20} & \cellcolor{gray!20} & \cellcolor{gray!20} & \cellcolor{gray!20} \\
     FLATTEN~\cite{cong2023flatten} \textit{\textcolor{gray}{{\fontsize{6}{12}\selectfont ICLR'24}}} & &  & O & \ding{51}  & \ding{51} &  &  &  & \ding{51} & \cellcolor{gray!20} & \cellcolor{gray!20} &  \cellcolor{gray!20} & \cellcolor{gray!20} & \cellcolor{gray!20} & \cellcolor{gray!20} & \cellcolor{gray!20} & \cellcolor{gray!20} \\
     Follow-Your-Pose~\cite{ma2024follow} \textit{\textcolor{gray}{{\fontsize{6}{12}\selectfont AAAI'24}}} &  &  & P &  &  &  & &  & \ding{51} & \cellcolor{gray!20} & \cellcolor{gray!20} & \cellcolor{gray!20} & \cellcolor{gray!20} &  \cellcolor{gray!20} & \cellcolor{gray!20} & \cellcolor{gray!20} & \cellcolor{gray!20} \\
     FreSCo~\cite{yang2024fresco} \textit{\textcolor{gray}{{\fontsize{6}{12}\selectfont CVPR'24}}} & &   & S & \ding{51} & &  &  &   & \ding{51} & \cellcolor{gray!20} & \cellcolor{gray!20} & \cellcolor{gray!20} & \cellcolor{gray!20} & \cellcolor{gray!20} & \cellcolor{gray!20} & \cellcolor{gray!20} & \cellcolor{gray!20} \\
     FlowVid~\cite{liang2024flowvid} \textit{\textcolor{gray}{{\fontsize{6}{12}\selectfont CVPR'24}}} & & & O+D  &    & \ding{51} &  &  &  & \ding{51} & \cellcolor{gray!20} & \cellcolor{gray!20} & \cellcolor{gray!20} &  \cellcolor{gray!20} & \cellcolor{gray!20} & \cellcolor{gray!20} & \cellcolor{gray!20} & \cellcolor{gray!20}  \\
     RAVE~\cite{kara2024rave} \textit{\textcolor{gray}{{\fontsize{6}{12}\selectfont CVPR'24}}} & & & D  &  & \ding{51} &  &  & & \ding{51} & \cellcolor{gray!20} & \cellcolor{gray!20} & \cellcolor{gray!20} & \cellcolor{gray!20} & \cellcolor{gray!20} & \cellcolor{gray!20} & \cellcolor{gray!20} & \cellcolor{gray!20} \\

     CoDeF~\cite{ouyang2024codef} \textit{\textcolor{gray}{{\fontsize{6}{12}\selectfont CVPR'24}}} & \ding{51} &  & &  & &   &  &  & \ding{51}  &\cellcolor{gray!20} & \cellcolor{gray!20} & \cellcolor{gray!20} & \cellcolor{gray!20} & \cellcolor{gray!20} & \cellcolor{gray!20} & \cellcolor{gray!20} & \cellcolor{gray!20} \\
     VMC~\cite{jeong2024vmc} \textit{\textcolor{gray}{{\fontsize{6}{12}\selectfont CVPR'24}}}& \ding{51}  &  & Mo &  & \ding{51}  &   & &  & \ding{51} & \cellcolor{gray!20} & \cellcolor{gray!20} & \cellcolor{gray!20} &  \cellcolor{gray!20} & \cellcolor{gray!20} & \cellcolor{gray!20} & \cellcolor{gray!20} & \cellcolor{gray!20} \\
     SimDA~\cite{xing2024simda} \textit{\textcolor{gray}{{\fontsize{6}{12}\selectfont CVPR'24}}}& &  & &   & \ding{51}  &   & &  & \ding{51} & \cellcolor{gray!20} & \cellcolor{gray!20} & \cellcolor{gray!20} &  \cellcolor{gray!20} & \cellcolor{gray!20} & \cellcolor{gray!20} & \cellcolor{gray!20} & \cellcolor{gray!20} \\
     LAMP~\cite{wu2023lamp} \textit{\textcolor{gray}{{\fontsize{6}{12}\selectfont CVPR'24}}} & \ding{51} & \ding{51} & C &  & \ding{51} &   &  &  & \ding{51} &\cellcolor{gray!20} & \cellcolor{gray!20} & \cellcolor{gray!20} & \cellcolor{gray!20} & \cellcolor{gray!20} & \cellcolor{gray!20} & \cellcolor{gray!20} & \cellcolor{gray!20} \\
     CusAV~\cite{ren2024customize} \textit{\textcolor{gray}{{\fontsize{6}{12}\selectfont ECCV'24}}} & \ding{51} &  & & &  &  & & & \ding{51}& \cellcolor{gray!20} &  \cellcolor{gray!20} & \cellcolor{gray!20} & \cellcolor{gray!20} & \cellcolor{gray!20} & \cellcolor{gray!20} & \cellcolor{gray!20} & \cellcolor{gray!20}\\
     MotionDirector~\cite{zhao2023motiondirector} \textit{\textcolor{gray}{{\fontsize{6}{12}\selectfont ECCV'24}}}& \ding{51} & &  & &  &  & &  & \ding{51} & \cellcolor{gray!20} & \cellcolor{gray!20} &  \cellcolor{gray!20} & \cellcolor{gray!20} & \cellcolor{gray!20} & \cellcolor{gray!20} & \cellcolor{gray!20} & \cellcolor{gray!20} \\
     NeRCan~\cite{chen2024narcan} \textit{\textcolor{gray}{{\fontsize{6}{12}\selectfont NeurIPS'24}}} & \ding{51} &  & &  & &   &  &  & \ding{51}  &\cellcolor{gray!20} & \cellcolor{gray!20} & \cellcolor{gray!20} &  \cellcolor{gray!20} & \cellcolor{gray!20} & \cellcolor{gray!20} & \cellcolor{gray!20} & \cellcolor{gray!20} \\
     \rowcolor{gray!20} \multicolumn{18}{|c|}{{\textit{Diverse Editing}}} \\

     Text2LIVE~\cite{bar2022text2live} \textit{\textcolor{gray}{{\fontsize{6}{12}\selectfont ECCV'22}}} & & \ding{51}  & A &   & & \ding{51} & \ding{51} & \ding{51} & \ding{51} & \ding{51}  &  & & & \ding{51}  &    & \ding{51}  & \ding{51} \\
     MoCA~\cite{yan2023motion} \textit{\textcolor{gray}{{\fontsize{6}{12}\selectfont arXiv'23}}} &  & \ding{51} & O & &  & \ding{51} & \ding{51} & \ding{51} & \ding{51} & \ding{51}  & \ding{51} &  & & \ding{51} &  \ding{51} & \ding{51} & \ding{51}\\
     DiffusionAtlas~\cite{chang2023diffusionatlas} \textit{\textcolor{gray}{{\fontsize{6}{12}\selectfont arXiv'23}}} & \ding{51} & \ding{51} & A &   & \ding{51} & \ding{51} & \ding{51} & \ding{51} & \ding{51} & \ding{51}  & &  & & \ding{51}  &  & \ding{51}  & \ding{51} \\
     Make-A-Prota.~\cite{zhao2023make} \textit{\textcolor{gray}{{\fontsize{6}{12}\selectfont arXiv'23}}} & \ding{51} &  & D+Ma & \ding{51} & \ding{51} & \ding{51} & \ding{51} & \ding{51} & \ding{51} & \ding{51} &  & & & \ding{51}  &  & \ding{51} & \ding{51}\\
     MagicEdit~\cite{liew2023magicedit} \textit{\textcolor{gray}{{\fontsize{6}{12}\selectfont arXiv'23}}} & & & D/P &   & \ding{51} & \ding{51} & \ding{51} &  & \ding{51} & \ding{51} &  & & & \ding{51} &  \ding{51} & \ding{51} & \ding{51}\\
     VidEdit~\cite{couairon2023videdit} \textit{\textcolor{gray}{{\fontsize{6}{12}\selectfont TMLR'23}}}& & \ding{51}  & A+H+Ma &  & \ding{51} & \ding{51} & \ding{51} & \ding{51} & \ding{51} & \ding{51}  &  & & & \ding{51}   &   & \ding{51}  & \ding{51} \\
     STL~\cite{lee2023shape} \textit{\textcolor{gray}{{\fontsize{6}{12}\selectfont CVPR'23}}} & & \ding{51}  & A+Ma  &  & & \ding{51} & \ding{51} & \ding{51} & \ding{51} & \ding{51}  &  & & & \ding{51}  &    & \ding{51}  & \ding{51} \\
     T2V-Zero~\cite{khachatryan2023text2video} \textit{\textcolor{gray}{{\fontsize{6}{12}\selectfont ICCV'23}}} & & \ding{51} & Ma &  &  & \ding{51} & \ding{51} &  & \ding{51} & \ding{51} &  & & & \ding{51} &  \ding{51} & \ding{51} & \ding{51}\\
     Tune-A-Video~\cite{wu2023tune} \textit{\textcolor{gray}{{\fontsize{6}{12}\selectfont ICCV'23}}}& \ding{51} &  & &   & \ding{51} & \ding{51} & \ding{51} & \ding{51} & \ding{51} & \ding{51}  &  & & \ding{51} & \ding{51} & \ding{51}  & \ding{51} & \ding{51}\\
     Gen-1~\cite{esser2023structure} \textit{\textcolor{gray}{{\fontsize{6}{12}\selectfont ICCV'23}}}  & &  & D+Ma &  & &  & \ding{51} &  & \ding{51} & &  & & &  &   & \ding{51} & \ding{51}\\
     StableVideo~\cite{chai2023stablevideo} \textit{\textcolor{gray}{{\fontsize{6}{12}\selectfont ICCV'23}}} & & \ding{51} & A+D+C &  & & \ding{51} & \ding{51} & \ding{51} & \ding{51} & \ding{51}  &  & & & \ding{51}  &    & \ding{51}  & \ding{51} \\
     VideoComposer~\cite{wang2024videocomposer} \textit{\textcolor{gray}{{\fontsize{6}{12}\selectfont NeurIPS'23}}} &  & & D/S/Ma/Mo/SI  &   &  &  \ding{51} & \ding{51} &  & \ding{51} & \ding{51} & \ding{51}  &  & & \ding{51} &   \ding{51} & \ding{51} & \ding{51}\\
     UniEdit~\cite{bai2024uniedit} \textit{\textcolor{gray}{{\fontsize{6}{12}\selectfont arXiv'24}}} &  &  & & \ding{51} & \ding{51} & \ding{51} & \ding{51}& \ding{51} & \ding{51} & \ding{51} & \ding{51} &  & \ding{51} & \ding{51} & \ding{51} & \ding{51} & \ding{51}\\
     AnyV2V~\cite{ku2024anyv2v} \textit{\textcolor{gray}{{\fontsize{6}{12}\selectfont arXiv'24}}} &  & \ding{51} & & \ding{51} & \ding{51} & \ding{51} & \ding{51} & \ding{51} & \ding{51} &  \ding{51} &  &  \ding{51} & & \ding{51} &   \ding{51} & \ding{51} & \ding{51}\\
     Edit-A-Video~\cite{shin2024edit} \textit{\textcolor{gray}{{\fontsize{6}{12}\selectfont ACML'24}}} & \ding{51} &  & & \ding{51} & \ding{51} & \ding{51} & \ding{51} & \ding{51} & \ding{51} & \ding{51} &  & & & \ding{51} &    & \ding{51} & \ding{51}\\
     TokenFlow~\cite{geyer2023tokenflow} \textit{\textcolor{gray}{{\fontsize{6}{12}\selectfont ICLR'24}}}& &  & & \ding{51} & \ding{51} & \ding{51} & \ding{51} & \ding{51}  & \ding{51} & \ding{51}  &  & &  & \ding{51} &   & \ding{51} & \ding{51}\\
     Ground-A-Video~\cite{jeong2023ground} \textit{\textcolor{gray}{{\fontsize{6}{12}\selectfont ICLR'24}}} & & & D+O+B &    & \ding{51} & \ding{51} & \ding{51} & \ding{51} & \ding{51} & \ding{51} &   &  & & \ding{51} &   & \ding{51} & \ding{51} \\
     ControlVideo~\cite{zhao2023controlvideo} \textit{\textcolor{gray}{{\fontsize{6}{12}\selectfont ICLR'24}}} & &  & C/H/D/P &  & \ding{51} & \ding{51} & \ding{51} & \ding{51} & \ding{51}  & \ding{51} & \ding{51} & & & \ding{51} &    & \ding{51} & \ding{51}\\
     CCEdit~\cite{feng2024ccedit} \textit{\textcolor{gray}{{\fontsize{6}{12}\selectfont CVPR'24}}} & & \ding{51} & D/S &  & & \ding{51} & \ding{51} & \ding{51} & \ding{51} & \ding{51} &  & & & \ding{51} &  \ding{51}  & \ding{51} & \ding{51}\\
     Fairy~\cite{wu2024fairy} \textit{\textcolor{gray}{{\fontsize{6}{12}\selectfont CVPR'24}}}&  & \ding{51} & & \ding{51} &  & \ding{51} & \ding{51} & \ding{51} & \ding{51} & \ding{51}  &  & & & \ding{51} &   \ding{51} & \ding{51} & \ding{51}\\
     Video-P2P~\cite{liu2024video} \textit{\textcolor{gray}{{\fontsize{6}{12}\selectfont CVPR'24}}}& \ding{51} &  & & \ding{51} & \ding{51} & \ding{51} & \ding{51} & \ding{51} & \ding{51} & \ding{51} & & & & \ding{51} &    & \ding{51} & \ding{51}\\
     MotionEditor~\cite{tu2024motioneditor} \textit{\textcolor{gray}{{\fontsize{6}{12}\selectfont CVPR'24}}} & &  & P+Ma &  & \ding{51}  & \ding{51} &  & & & & \ding{51}    & & &  &   & & \\
     VidToMe~\cite{li2024vidtome} \textit{\textcolor{gray}{{\fontsize{6}{12}\selectfont CVPR'24}}}&  &  & & \ding{51} & \ding{51} & \ding{51} & \ding{51} & \ding{51} & \ding{51} & \ding{51} &  & & & \ding{51} &   & \ding{51} & \ding{51}\\
     SAVE~\cite{song2023save} \textit{\textcolor{gray}{{\fontsize{6}{12}\selectfont ECCV'24}}} & \ding{51} &  & Ma & &  & \ding{51} & &  &  & \ding{51} &  & & & \ding{51} &  &   & \\
     EVE~\cite{singer2024video} \textit{\textcolor{gray}{{\fontsize{6}{12}\selectfont ECCV'24}}} &  &  &  & &  & \ding{51} & \ding{51} &  & \ding{51} & \ding{51} &  & \ding{51} & \ding{51} & \ding{51} &   \ding{51} & \ding{51} & \ding{51}\\
     WAVE~\cite{fengwave} \textit{\textcolor{gray}{{\fontsize{6}{12}\selectfont ECCV'24}}} &  &  & O+Ma & \ding{51} & \ding{51} & \ding{51} & & \ding{51} & \ding{51} & \ding{51} &  & & & \ding{51} &   \ding{51} & \ding{51} & \\
     DeCo~\cite{zhong2024deco} \textit{\textcolor{gray}{{\fontsize{6}{12}\selectfont ECCV'24}}} & \ding{51} &  & D+A+P & &  & \ding{51} & \ding{51} & \ding{51} &  & \ding{51} &  & & & \ding{51} &  \ding{51} & \ding{51} & \ding{51}\\
     Videoshop~\cite{fan2024videoshop} \textit{\textcolor{gray}{{\fontsize{6}{12}\selectfont ECCV'24}}} &  & \ding{51} & Ma & & \ding{51} & \ding{51} & \ding{51} & \ding{51} &  & \ding{51} &  & \ding{51} & \ding{51} & \ding{51} &    &  & \ding{51}\\
     OCD~\cite{kahatapitiya2024object} \textit{\textcolor{gray}{{\fontsize{6}{12}\selectfont ECCV'24}}} &  &  & Ma & \ding{51} & \ding{51} & \ding{51} & &  & \ding{51} & \ding{51} &  & & & \ding{51} &    &  & \\
     DreamMotion~\cite{jeong2024dreammotion} \textit{\textcolor{gray}{{\fontsize{6}{12}\selectfont ECCV'24}}} &  &  & Ma & & \ding{51} & \ding{51} & & \ding{51} & \ding{51} & \ding{51} &  & & & \ding{51} &    & \ding{51} & \\
     MagDiff~\cite{zhaomagdiff} \textit{\textcolor{gray}{{\fontsize{6}{12}\selectfont ECCV'24}}} &  &  & Ma & &  & \ding{51} & \ding{51} &  & \ding{51} & \ding{51} & & & & \ding{51} &    & \ding{51} & \ding{51}\\
     COVE~\cite{wang2024cove} \textit{\textcolor{gray}{{\fontsize{6}{12}\selectfont NeurIPS'24}}} &  &  &  & \ding{51} & \ding{51} & \ding{51} & \ding{51} & \ding{51} & \ding{51} & \ding{51} &  &  & & \ding{51} &   \ding{51} & \ding{51} & \ding{51}\\
     \hline
     \rowcolor{cyan!10}
     $+$ \textbf{\texttt{UES} (Ours)} &  &  & &  & & \ding{51}  & \ding{51} & \ding{51} & \ding{51} & \ding{51} & \ding{51} & \ding{51} & \ding{51} & \ding{51} & \ding{51} &  \ding{51} & \ding{51}\\
     \hlineB{2.5}
   \end{tabular}
  \label{app_tab:review}
  \vspace{0.4cm}
\end{table*} 
\endgroup

\vspace{-0.8em}
\subsubsection*{Additional Control.} We categorize the additional control in video editing into three main types: \ding{182} \textit{Condition} (Cond.): Widely used to constrain the structure of the edited video, this approach incorporates elements, \textit{e.g.}, optical flow, depth maps, masks, edges, etc.; \ding{183} \textit{Attention Features} (Feat.): This enhances control during inference by injecting hidden features from a reference video to guide the structure of the edited video. It can be divided into two types: inversion-based features and motion-based features~\cite{sun2024diffusion}; \ding{184} \textit{DDIM Inversion} (DDIM): In this technique, the inverted noise from a reference video is progressively denoised during inference, ensuring consistency between the edited and reference videos. While these additional control strategies have significantly advanced controllable video generation/editing, it is important to acknowledge that the introduction of explicit controls can impose certain limitations on the model's flexibility in editing scenarios.

\vspace{-0.8em}
\subsubsection*{Editing Type.} Following~\cite{sun2024diffusion}, we classify video editing into four distinct types, as shown at the top of Figure~\ref{fig:example} in the main content. Foreground (Fore.) and Background (Back.) editing refer to modifying only specific parts of the foreground or background while keeping other regions as unchanged as possible. Overall/Style editing involves altering the entire video. It is worth noting that even when certain methods target specific regions, they often result in broader changes to the entire video. Thus we classify these models as overall editing as well. Composite (Comp.) editing requires more complex modifications, such as editing both the foreground and style while keeping the background as consistent as possible.

\vspace{-0.8em}
\subsubsection*{Editing Scenario.}
To further evaluate the model’s performance, we categorized the video editing scenarios into three main groups: \ding{182} \textit{Human/Animal}: 1) Appearance (App.) editing involves modifying the physical appearance of a person or animal, classified as foreground editing. 2) Motion (Mo.) or Pose editing focuses on action changes, such as altering actions from ``walking” to ``running.”; \ding{183} \textit{Object}: 1) Addition (Add.) and Removal (Rem.) editing refer to inserting or deleting specific objects in the video. 2) Replacement (Rep.) editing involves substituting one object for another, also classified under foreground editing; \ding{184} \textit{Environment}: 1) Weather (Wea.) editing involves composite editing, such as adding rainfall to a reference video featuring a seascape. This task not only requires adding raindrop effects but also capturing ripples in the water. 2) Time editing is a form of overall editing, adjusting the video based on prompts related to different seasons or times of day, while maintaining the original video style (\textit{e.g.}, a realistic video should not be transformed into an animated style). 3) Background (Back.) editing aligns with the typical background editing type. 

These eight categories are designed to assess various application scenarios for text-based video editing models. While this list may not be exhaustive, we plan to further expand it in future work.

\vspace{-0.8em}
\subsubsection*{Video-Video Pair \textit{vs.} Text-Video Pair-based.} While text-video pair datasets are widely used for text-to-video generation and editing, two works~\cite{cheng2023consistent, qin2023instructvid2vid} utilize video-video data for training. These datasets consist of a reference video, a human-instruction-like text prompt, and an edited video. Although this approach has seen some success at the image level~\cite{brooks2023instructpix2pix}, video datasets present greater challenges in terms of resource consumption and quality, since they rely on off-the-shelf video or image editing models (\textit{e.g.}, TAV~\cite{wu2023tune} and P2P~\cite{hertz2022prompt}) to create video pairs. While these works do not employ additional explicit structural control, their reliance on tightly paired training videos and the limited information provided by the text prompts further constrain the performance of their models.

\section{OmniBench-99}
\label{app_bench}

\subsection{Overview}
To comprehensively evaluate the performance of generative text-guided video editing models, we introduce OmniBench-99, a benchmark that enables both editing types and scenarios evaluation. Our dataset consists of 99 high-quality, open-license videos across three categories: \textit{Human/Animal}, \textit{Environment}, and \textit{Object}. For each video, we first create four editing-type prompts, \textit{i.e.}, Foreground, Background, Composite, and Style/Overall. Then, based on the video’s category, we generate specific editing-scenario prompts. 

\begin{figure}[t]
    \centering
   \includegraphics[width=1\linewidth]{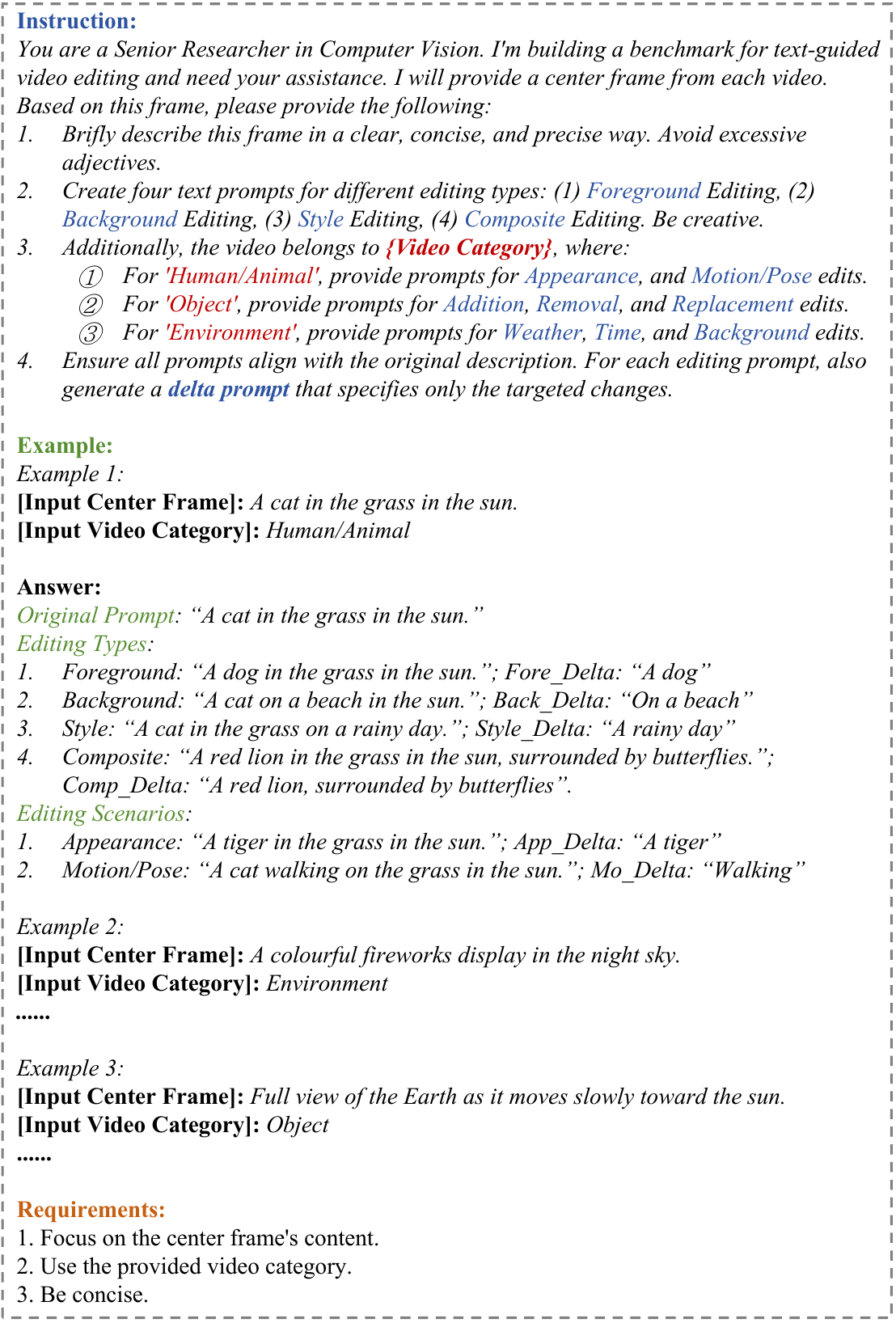}
   \vspace{-1.8em}
    \captionof{figure}{\label{fig:app_prompt}
    Text Prompt Demonstration. }
\end{figure}

To accommodate the variety of prompts accepted by existing models, we further design two types of text prompt, \textit{i.e.}, \textit{full description} and \textit{delta caption}.
To reduce the burden of manual data annotation, we leverage GPT-4V(ision)~\cite{openai2024gpt4vsystemcard} to automatically generate these prompts following the instructions shown in Figure~\ref{fig:app_prompt}.

\begin{figure*}[t]
    \centering
   \includegraphics[width=0.76\linewidth]{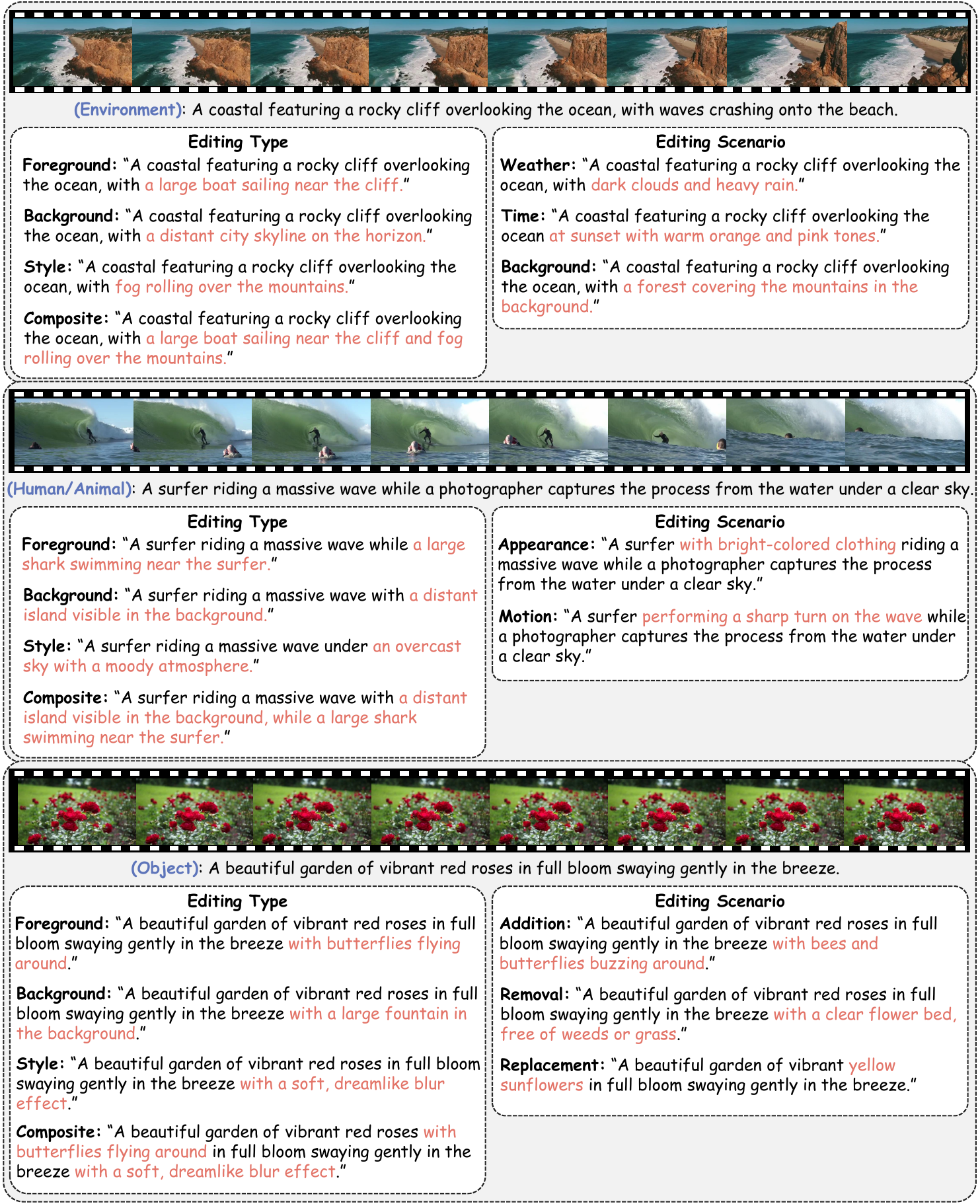}
   \vspace{-0.4em}
    \captionof{figure}{\label{fig:bench_example}
    Examples of OmniBench-99 benchmark. The delta prompts are highlighted in yellow-brown.}
\end{figure*}

\subsection{Human Inspection}

Human inspection plays a pivotal role in our OmniBench-99 since LLMs often experience some hallucinations. In video editing, while the range of potential edits is vast, we aim for reasonable and appropriate prompts. During the refinement process, we focus on ensuring that the prompt aligns with the video content and whether the resulting edited video adheres to \textit{real-world physics}. For example, consider a video of a car driving on a road with the prompt to change the road into a cracked canyon. Editing only the road would be ideal, but driving the car across a canyon is unrealistic. Ensuring such physical consistency is a key aspect of our approach:
\begingroup
\setlength{\tabcolsep}{3.4pt}
\begin{table*}[t]
\renewcommand{\arraystretch}{1.2}
  \centering
  \caption{More comparison with T(+I)2V generation models on full-dimension VBench.}
  \centering
  \vspace{-0.8em}
  \scriptsize
   \begin{tabular}{lccccccccccccccccccc}
     \hlineB{2.5}
     \textbf{Method} & \textbf{Total} & \textbf{QS} & \textbf{SS} & \textbf{SC} & \textbf{BC} & \textbf{TF} & \textbf{MS} & \textbf{DD} & \textbf{AQ} & \textbf{IQ} & \textbf{OC} & \textbf{MO} & \textbf{HA} & \textbf{Color} & \textbf{SR} & \textbf{Scene} & \textbf{AS} & \textbf{TS} & \textbf{OC}\\
     \hlineB{1.5}
     VideoCrafter2~\cite{chen2024videocrafter2} & 80.44 & 82.20 & 73.42 & \underline{96.85} & 98.22 & \underline{98.41} & 97.73 & \underline{42.50} & \textbf{63.13} & 67.22 & 92.55 & 40.66 & \textbf{95.00} & \underline{92.92} & 35.86 & 55.29 & \textbf{25.13} & 25.84 & 28.23 \\
     $+$ Fully Fine-tune & \underline{80.81} & \underline{82.57} & \underline{73.78} & \textbf{96.89} & 98.29 & 98.19 & 98.34 & \textbf{42.58} & 62.76 & \underline{68.33} & 92.41 & 42.11 & \underline{94.34} & \textbf{93.23} & 35.86 & 54.12 & \underline{25.02} & \textbf{26.92} & \underline{28.66}  \\
     $+$ LoRA Fine-tune & 80.49 & 82.29 & 73.30 & 96.13 & \underline{98.36} & 97.90 & \underline{98.45} & 42.32 & 62.82 & 67.78 & \underline{92.65} & \underline{44.34} & 93.12 & 91.46 & \underline{36.28} & \underline{55.87} & 24.87 & 25.48 & 27.98 \\
     \rowcolor{cyan!10}
     $+$ \textbf{\texttt{UES} (Ours)} & \textbf{81.90} & \textbf{83.51} & \textbf{75.45} & 96.44 & \textbf{98.76} & \textbf{99.32} & \textbf{98.87} & 42.40 & \underline{62.89} & \textbf{69.45} & \textbf{93.12} & \textbf{53.43} & 93.20 & 91.68 & \textbf{39.08} & \textbf{56.12} & 24.98 & \underline{26.71} & \textbf{28.93}\\
     \hline
     DynamiCrafter~\cite{xing2023dynamicrafter} & 80.60 & 82.74 & 72.09 & 93.81 & \underline{96.64} & \underline{98.24} & 96.84 & \underline{69.67} & 60.88 & 68.60 & 91.68 & 38.98 & \textbf{94.56} & \textbf{91.68} & 36.56 & 56.24 & \underline{23.12} & 26.71 & 26.44\\
     $+$ Fully Fine-tune & 80.42 & 82.52 & 72.03 & \underline{93.92} & 96.24 & 97.36 & 97.05 & 69.50 & \textbf{61.20} & \underline{69.00} & 90.75 & 39.50 & \underline{94.40} & \underline{91.50} & 36.80 & 56.50 & \textbf{23.30} & \underline{27.00} & 25.80\\
     $+$ LoRA Fine-tune & \underline{80.78} & \underline{82.89} & \underline{72.32} & 93.65 & \textbf{96.80} & 98.10 & \underline{97.20} & 69.40 & \underline{60.95} & 68.80 & \underline{91.85} & \underline{40.20} & 94.00 & 91.20 & \underline{37.10} & \underline{56.80} & 23.00 & 26.90 & \underline{26.60} \\
     \rowcolor{cyan!10}
     $+$ \textbf{\texttt{UES} (Ours)} & \textbf{81.73} & \textbf{83.71} & \textbf{73.76} & \textbf{94.10} & 96.66 & \textbf{98.90} & \textbf{97.50} & \textbf{69.80} & 60.88 & \textbf{70.50} & \textbf{92.30} & \textbf{44.50} & 93.80 & 91.32 & \textbf{39.50} & \textbf{57.20} & \underline{23.12} & \textbf{27.50} & \textbf{27.80} \\
     \hlineB{2.5}
   \end{tabular}
   \parbox{\textwidth}{
{\scriptsize \textbf{Note:} QS: Quality Score, SS: Semantic Score, SC: Subject Consistency, BC: Background Consistency, TF: Temporal Flickering, MS: Motion Smoothness, DD: Dynamic Degree, AQ: Aesthetic Quality, IQ: Imaging Quality, OC: Object Class, MO: Multiple Objects, HA: Human Action, SR: Spatial Relationship, AS: Appearance Style, TS: Temporal Style, OC: Overall Consistency.}
}
  \label{app_tab:quantitative_vbench}
  \vspace{-0.3cm}
\end{table*} 
\endgroup

\begin{itemize}[leftmargin=*]
    \vspace{0.4em}
    \item \textbf{Editing Type} 
    \vspace{0.2em}
    \begin{itemize}[leftmargin=*]
    \item\textit{Foreground}: Ensure that the prompt clearly specifies the foreground object and that the edits (\textit{e.g.}, changes in color, shape, or pose) are both reasonable and actionable. The foreground should be distinct from the background and other video elements to avoid ambiguity. 
    \vspace{0.1em}
    \item\textit{Background}: The prompt should clearly suggest modifications to the background, such as changing the setting, adjusting depth of field, or altering the environment's style. Verify that the target for editing is appropriate, such as focusing on a natural background rather than the primary subject.
    \vspace{0.1em}
    \item\textit{Style}: Ensure that the prompt proposes changes to the video’s overall or localized style, such as color tone, filters, or texture, and that these changes are consistent with the video content.
    \vspace{0.1em}
    \item\textit{Composite}: The prompt should involve multiple (at least two) elements, ensuring the edits work together cohesively without conflict. Check if all parts of the composite edit are feasible and correspond to identifiable points in the video.
    \end{itemize}
    \vspace{0.4em}
    \item \textbf{Editing Scenario}
    \vspace{0.2em}
    \begin{itemize}[leftmargin=*]
    \vspace{0.1em}
    \item Human/Animal
    \vspace{0.1em}
    \begin{itemize}[leftmargin=*]
    \item[\ding{172}]\textit{Appearance}: Evaluate whether the prompt clearly specifies the visual characteristics of a person or animal, such as clothing, skin tone, or hairstyle. Assess whether the requested edits are appropriate and relevant to the main subject of the video.
    \vspace{0.1em}
    \item[\ding{173}]\textit{Motion/Pose}: Ensure the prompt addresses the movement within the video. The description should be clear, and the requested motion changes should maintain the flow and coherence of the video.
    \end{itemize}
    \vspace{0.1em}
    \item Object 
    \vspace{0.1em}
    \begin{itemize}[leftmargin=*]
    \item[\ding{172}]\textit{Addition}: Ensure the prompt clearly specifies the object or element to be added and that it integrates naturally into the existing scene.
    \vspace{0.1em}
    \item[\ding{173}]\textit{Removal}: Verify that the prompt's request for object removal aligns with the actual video content and that the removal does not disrupt the scene's continuity.
    \vspace{0.1em}
    \item[\ding{174}]\textit{Replacement}: Check that the prompt provides clear instructions for object replacement and that the new object is appropriate for the video's context.
    \end{itemize}
    \vspace{0.1em}
    \item Environment
    \vspace{0.1em}
    \begin{itemize}[leftmargin=*]
    \item[\ding{172}]\textit{Weather}: Ensure the prompt provides clear editing instructions for weather elements, such as rain, sunshine, or snow, and that these changes are realistically reflected in the video’s background.
    \vspace{0.1em}
    \item[\ding{173}]\textit{Time}: Verify whether the prompt specifies time-related scene transitions, like day turning to night or seasonal shifts, and assess whether these changes are implemented logically.
    \vspace{0.1em}
    \item[\ding{174}]\textit{Background}: Ensure the prompt suggests appropriate modifications to the video’s background, such as switching urban settings to natural landscapes, while maintaining consistency with the video’s main theme.
    \end{itemize}
    \end{itemize}
\end{itemize}
We provide examples of OmniBench-99 in Figure~\ref{fig:bench_example}. The full benchmark will be released with our code in June.







\begingroup
\setlength{\tabcolsep}{2pt}
\begin{table}[t]
\renewcommand{\arraystretch}{1.1}
  \centering
  \caption{More comparison on LOVEU-TGVE-2023~\cite{wu2023cvpr} and BalanceCC~\cite{feng2024ccedit}, which only focus on editing-type evaluations.}
  \centering
  \vspace{-0.8em}
  \scriptsize
   \begin{tabular}{l|cc|cc}
     \hlineB{2.5}
     \multirow{2}{*}{\textbf{Method}} & \multicolumn{2}{c}{\textbf{LOVEU}} & \multicolumn{2}{c}{\textbf{BalanceCC}}\\
     \cline{2-3}\cline{4-5}
     & \textbf{CLIP Frame$\uparrow$} & \textbf{PickScore$\uparrow$} & \textbf{CLIP Frame$\uparrow$} & \textbf{PickScore$\uparrow$} \\
     \hlineB{2}
     ControlVideo~\cite{zhao2023controlvideo} & 0.930 & 0.201 & 0.950 & 0.210\\
     Tune-A-Video~\cite{wu2023tune}  & 0.924 & 0.204 & 0.937 & 0.206\\
    Pix2Video~\cite{ceylan2023pix2video}  & 0.916 & 0.201 & 0.939 & 0.208\\
     RAV~\cite{yang2023rerender}  & 0.909 & 0.196 & 0.928 & 0.201\\
     TokenFlow~\cite{geyer2023tokenflow}  & 0.940 & 0.205 & 0.949 & 0.210\\
     InsV2V~\cite{cheng2023consistent}  & 0.911 & 0.208 & - & -\\
     Video-P2P~\cite{liu2024video}  & 0.935 & 0.201 & - & - \\
     CCEdit~\cite{feng2024ccedit}  & - & - & 0.936 & 0.213\\
     \hline
     \rowcolor{lightgray!20}
     VideoCrafter2~\cite{chen2024videocrafter2} & \multicolumn{4}{c}{{\textit{No Editing Capability}}}\\
     \rowcolor{cyan!10}
     $+$ \textbf{\texttt{UES} (Ours)}  & \textbf{0.960} & \underline{0.208} & \underline{0.959} & \textbf{0.216} \\
     \hline
     \rowcolor{lightgray!20}
     DynamiCrafter~\cite{xing2023dynamicrafter} & \multicolumn{4}{c}{{\textit{No Editing Capability}}}\\
     \rowcolor{cyan!10}
     $+$ \textbf{\texttt{UES} (Ours)} &  \underline{0.958} & \textbf{0.209} &  \textbf{0.963} & \underline{0.214}\\
     \hlineB{2.5}
   \end{tabular}
  \label{app_tab:quan}
\end{table} 
\endgroup

\section{More Experimental Results}
\label{app_details}

\subsection{Video Generation Evaluation}
\label{app_more_gen}

\subsubsection*{More Comparison on VBench.} To more comprehensively evaluate UES, we conduct tests across all dimensions of VBench \cite{huang2023vbench}. The results, shown in Table~\ref{app_tab:quantitative_vbench}, highlight \textbf{\texttt{UES}}'s superiority in enhancing video generation while simultaneously unlocking universal editing capabilities.

\vspace{-1em}
\subsubsection*{More T2V Comparisons.} 
\begin{figure*}[t]
    \centering
   \includegraphics[width=0.74\linewidth]{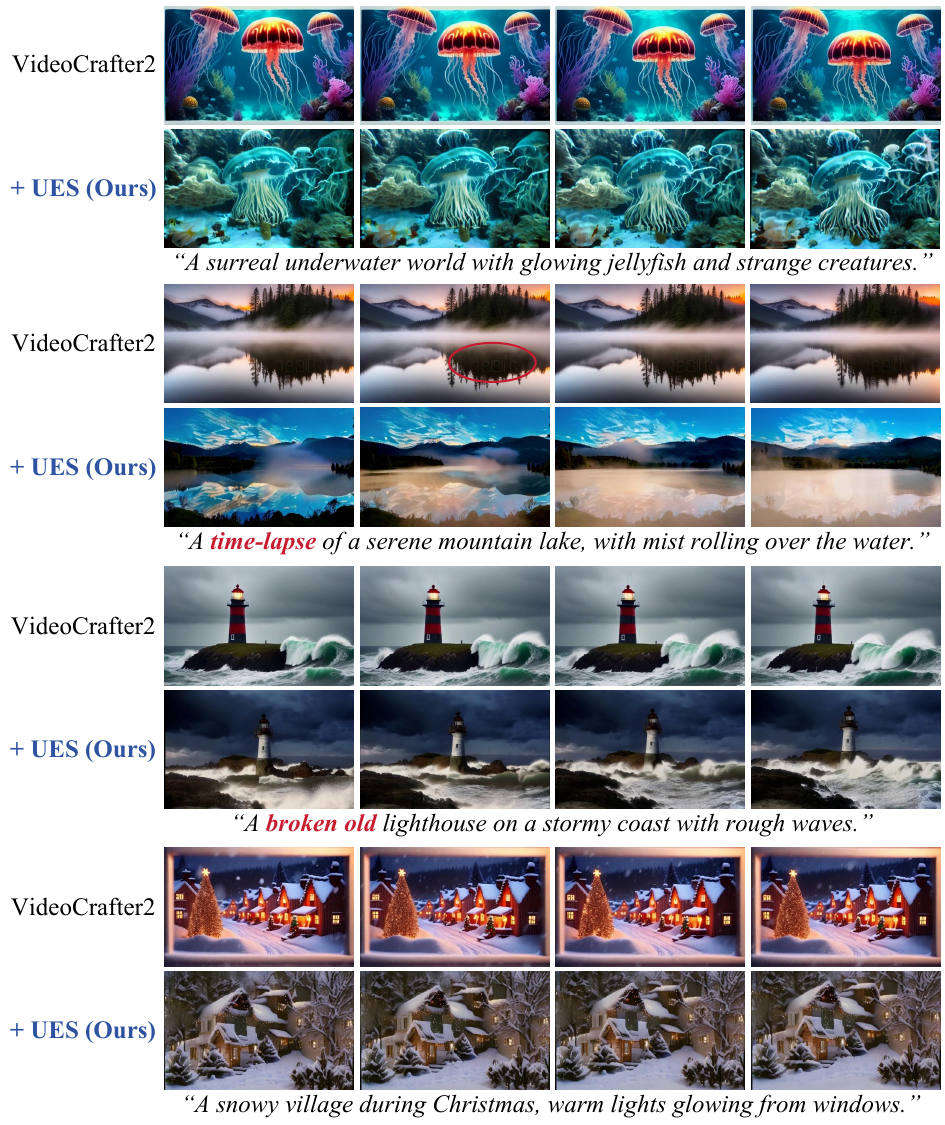}
   \vspace{-0.4em}
    \captionof{figure}{\label{fig:app_vc_vidgen}
    More video generation results of UES on VideoCrafter2.}
\end{figure*}

\begin{figure*}[t]
    \centering
   \includegraphics[width=0.7\linewidth]{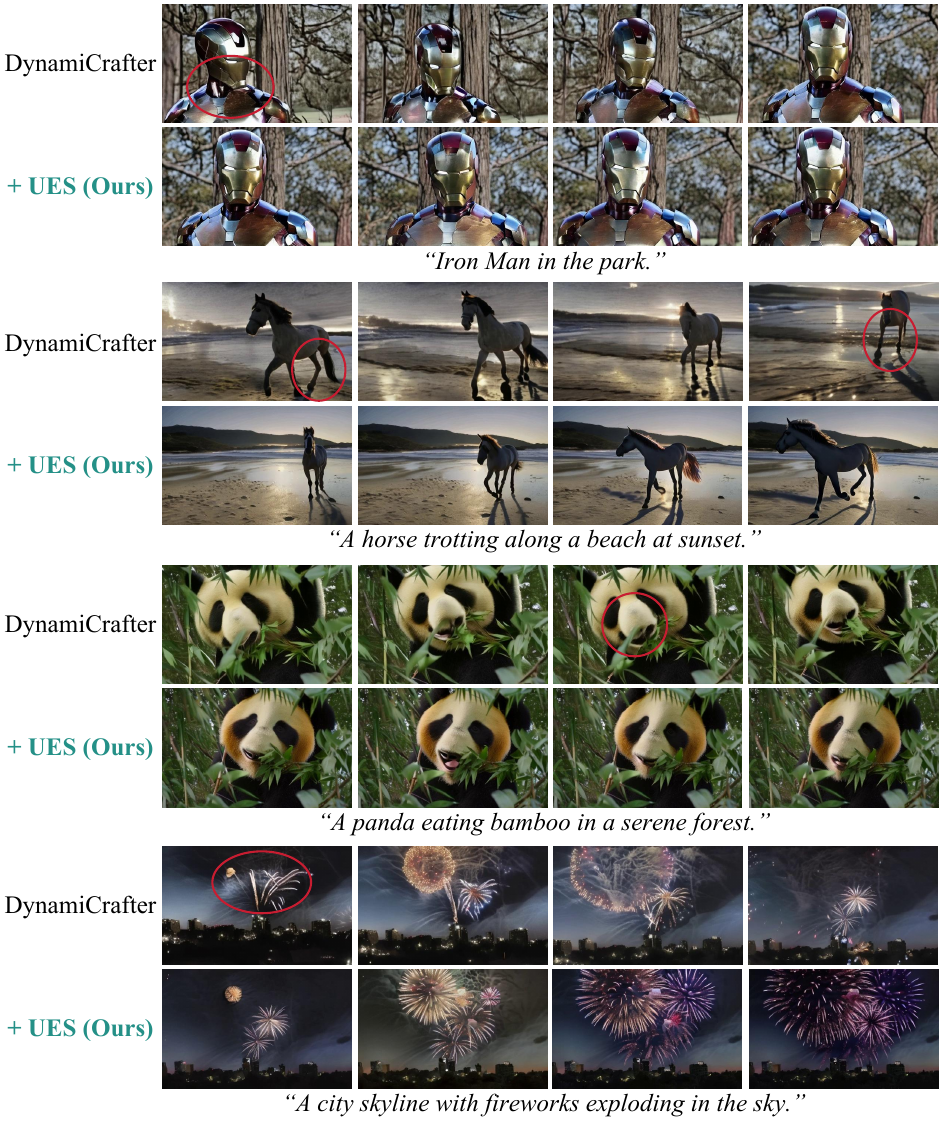}
   \vspace{-0.4em}
    \captionof{figure}{\label{fig:app_dy_vidgen}
    More video generation results of UES on DynamiCrafter.}
    \vspace{-0.4em}
\end{figure*}

More comparison of UES-enhanced generation performance on VideoCrafter2 and DynamiCrafter are presented in Figure~\ref{fig:app_vc_vidgen} and Figure~\ref{fig:app_dy_vidgen}, respectively. These results demonstrate the significant improvements achieved by UES in text alignment, temporal consistency, and overall realism.

\subsection{Video Editing Evaluation}
\label{app_more_edit}

\subsubsection*{More Comparison on LOVEU and BalanceCC.}
To further evaluate the effectiveness of \textbf{\texttt{UES}}, we conducted additional quantitative comparisons against state-of-the-art methods using popular benchmarks: LOVEU-TGVE-2023 \cite{wu2023cvpr} and BalanceCC \cite{feng2024ccedit}. As shown in Table~\ref{app_tab:quan}, models enhanced with UES consistently outperform baseline models.

Since these two benchmarks only focus on evaluating editing \textit{types}, we encourage future research to further evaluate editing \textit{scenarios} using our proposed OmniBench-99 benchmark (see Table~\ref{tab:quantitative} for experiments). This will allow for a more comprehensive assessment of model performance and generalizability.

\vspace{-1em}
\subsubsection*{More Video Editing Comparisons.}
\begin{figure*}[t]
    \centering
    \vspace{-0.8em}
   \includegraphics[width=0.75\linewidth]{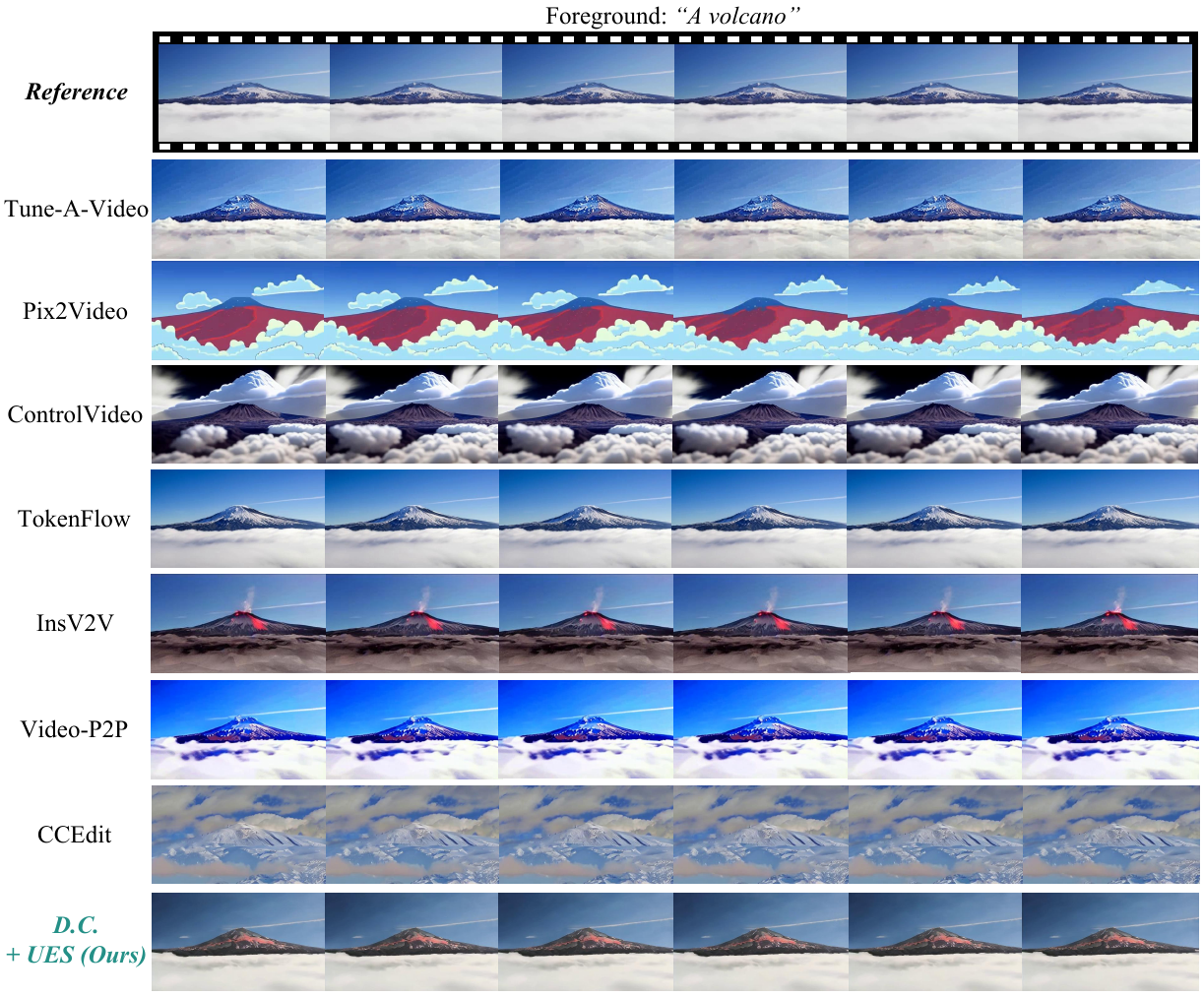}
   \vspace{-0.8em}
    \captionof{figure}{\label{fig:suppl_editing_type_fore}
    Video editing comparison: \textit{Editing-Type-Foreground}.}
    \vspace{-0.4em}
\end{figure*}

\begin{figure*}[t]
    \centering
    \vspace{-0.4em}
   \includegraphics[width=0.75\linewidth]{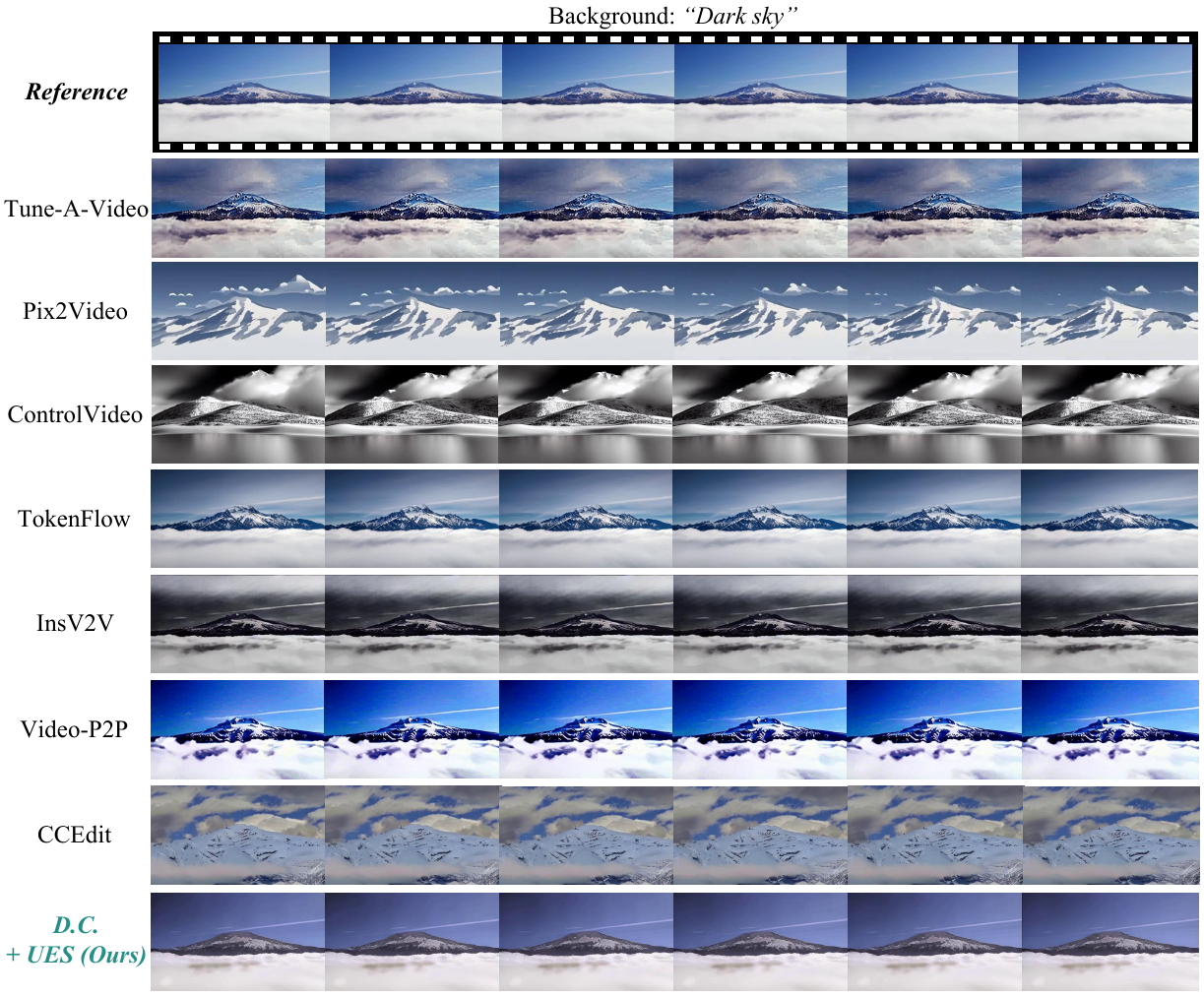}
   \vspace{-0.8em}
    \captionof{figure}{\label{fig:suppl_editing_type_back}
    Video editing comparison: \textit{Editing-Type-Background}.}
    \vspace{-0.8em}
\end{figure*}

\begin{figure*}[t]
    \centering
    \vspace{-0.8em}
   \includegraphics[width=0.75\linewidth]{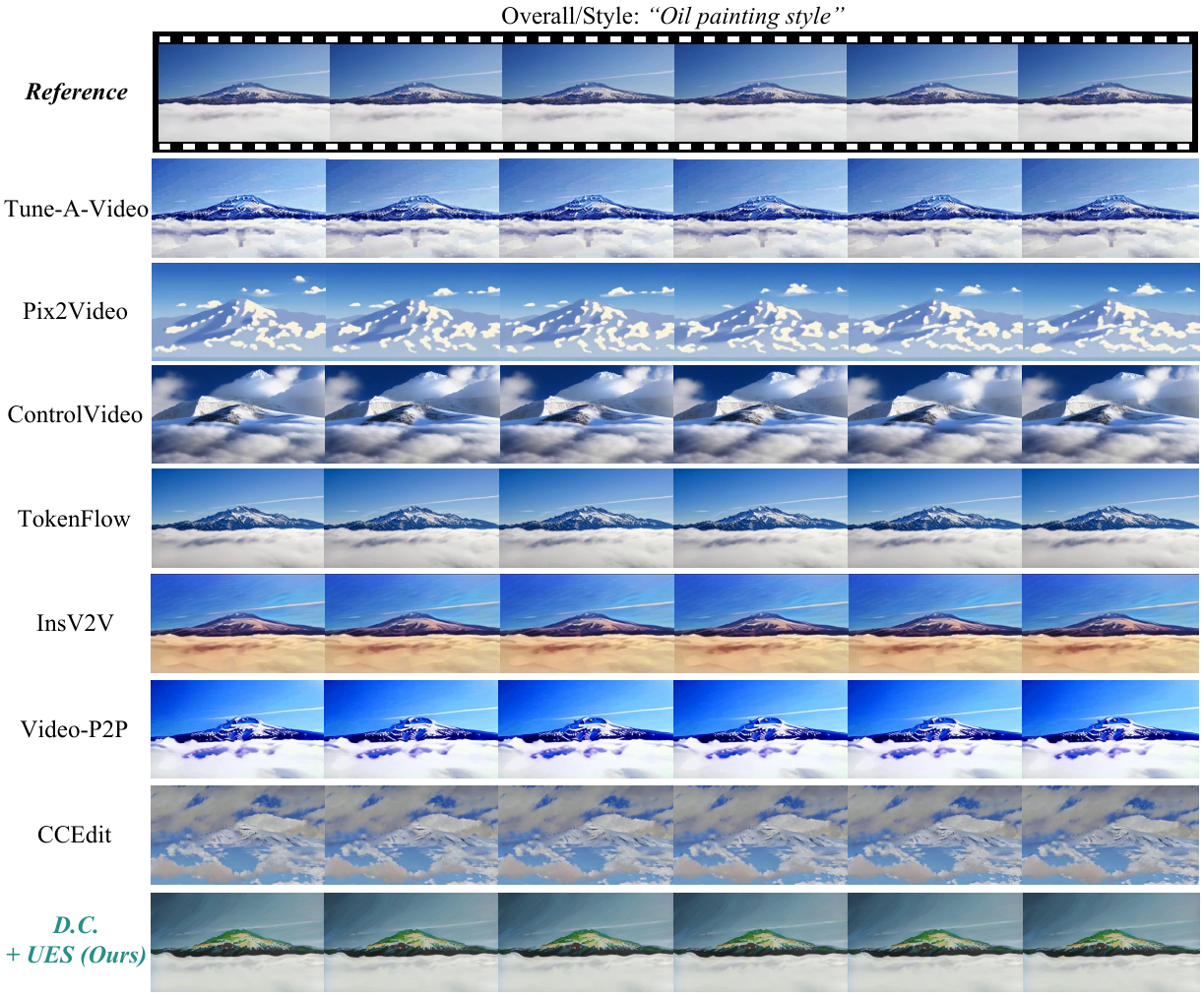}
   \vspace{-0.8em}
    \captionof{figure}{\label{fig:suppl_editing_type_style}
    Video editing comparison: \textit{Editing-Type-Style}.}
    \vspace{-0.4em}
\end{figure*}

\begin{figure*}[t]
    \centering
    \vspace{-0.4em}
   \includegraphics[width=0.75\linewidth]{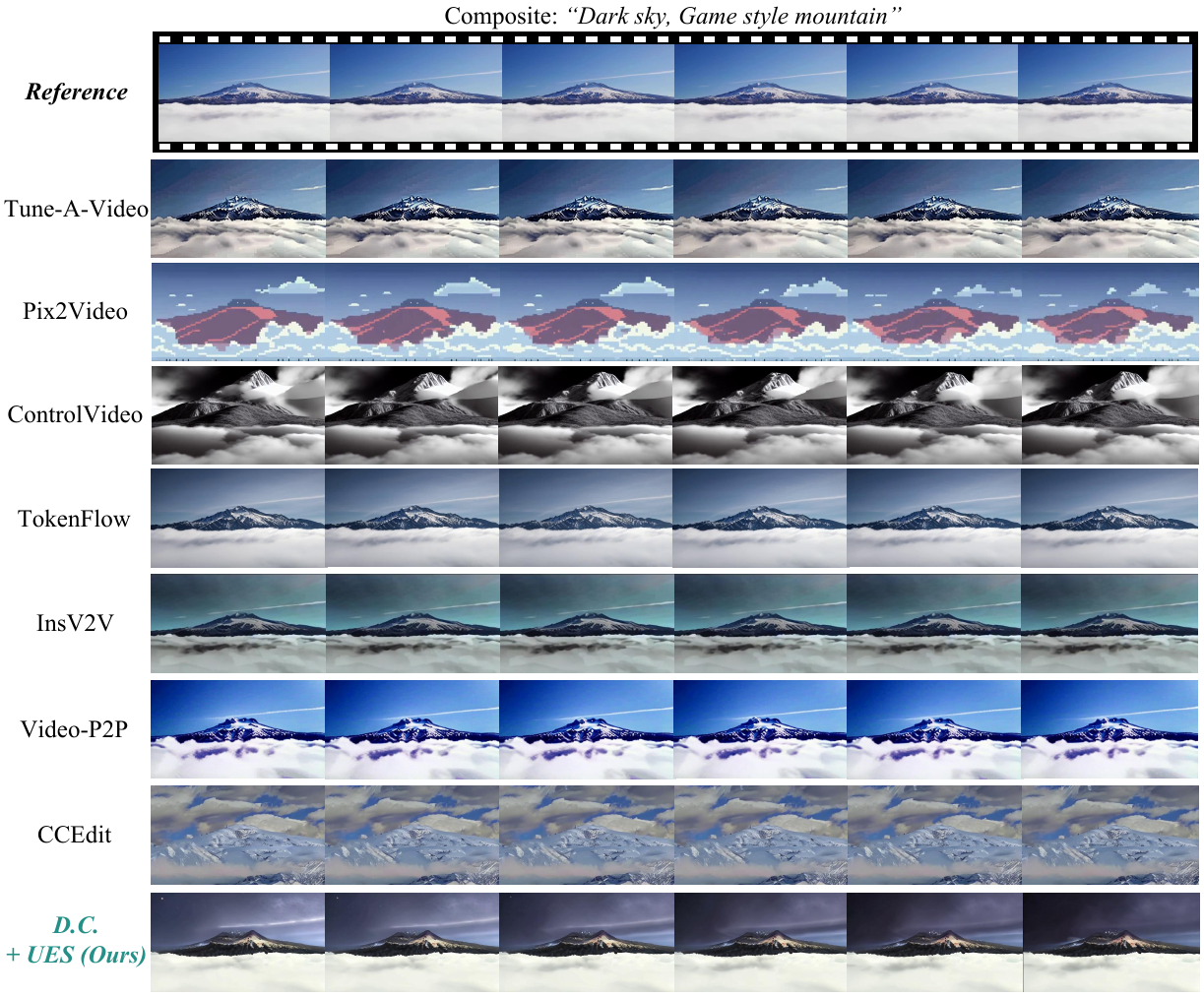}
   \vspace{-0.8em}
    \captionof{figure}{\label{fig:suppl_editing_type_compo}
    Video editing comparison: \textit{Editing-Type-Composite}.}
    \vspace{-0.8em}
\end{figure*}

\begin{figure*}[t]
    \centering
    \vspace{-0.8em}
   \includegraphics[width=0.75\linewidth]{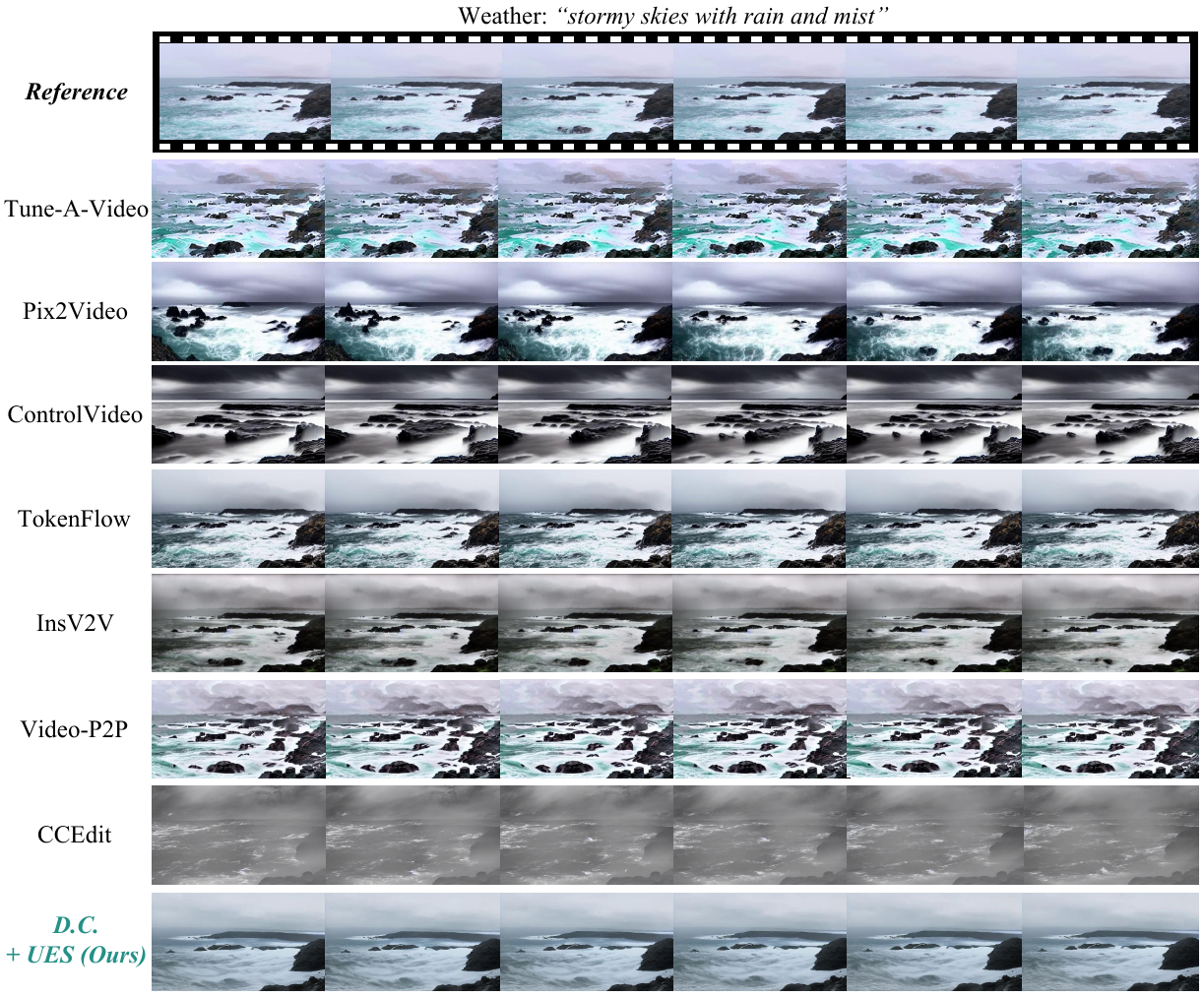}
   \vspace{-0.8em}
    \captionof{figure}{\label{fig:suppl_editing_env_weather}
    Video editing comparison: \textit{Editing-Scenario-Environment-Weather}.}
    \vspace{-0.4em}
\end{figure*}

\begin{figure*}[t]
    \centering
    \vspace{-0.4em}
   \includegraphics[width=0.75\linewidth]{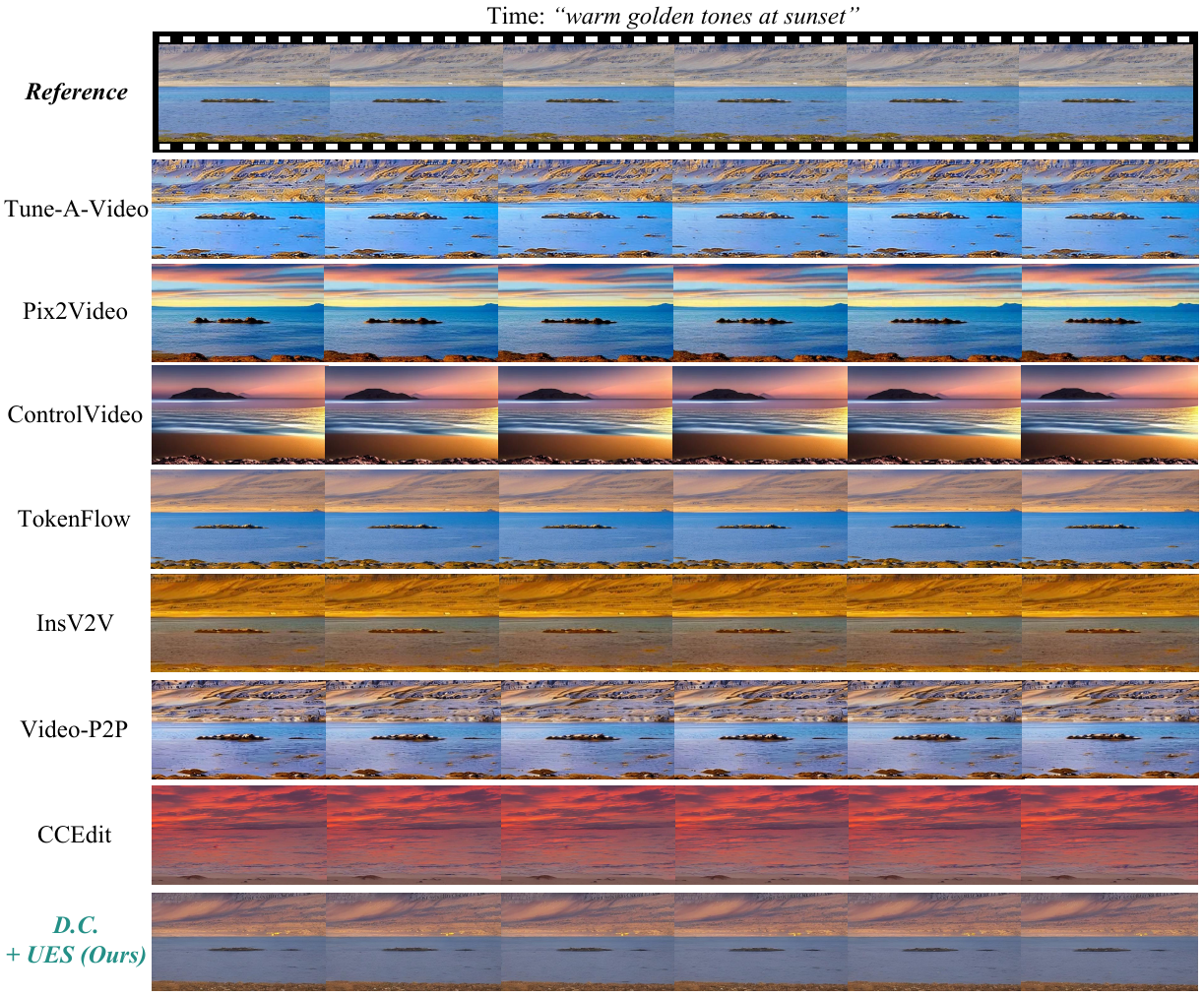}
   \vspace{-0.8em}
    \captionof{figure}{\label{fig:suppl_editing_env_time}
    Video editing comparison: \textit{Editing-Scenario-Environment-Time}.}
    \vspace{-0.8em}
\end{figure*}

\begin{figure*}[t]
    \centering
    \vspace{-0.8em}
   \includegraphics[width=0.75\linewidth]{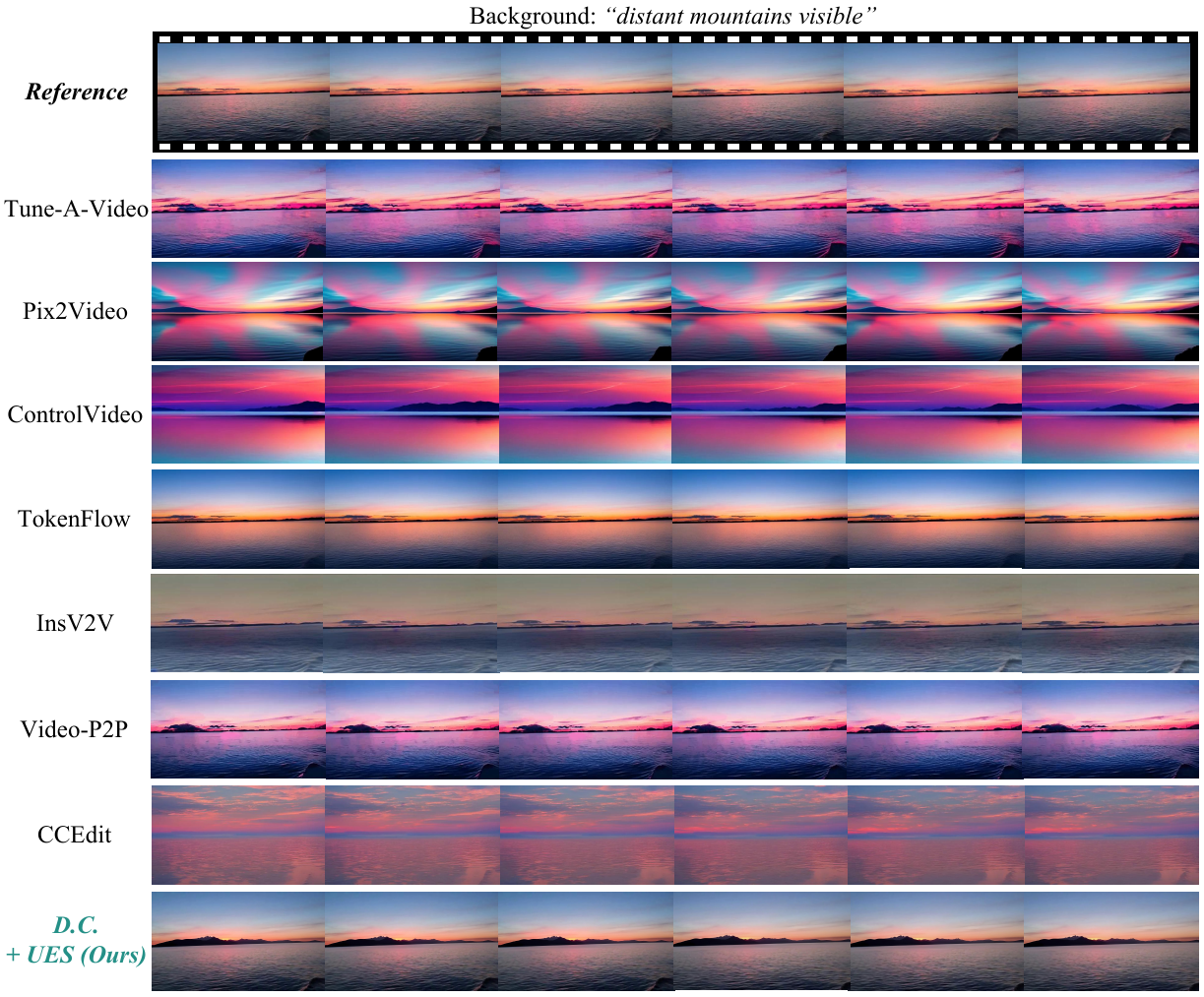}
   \vspace{-0.8em}
    \captionof{figure}{\label{fig:suppl_editing_env_back}
    Video editing comparison: \textit{Editing-Scenario-Environment-Background}.}
    \vspace{-0.4em}
\end{figure*}

\begin{figure*}[t]
    \centering
    \vspace{-0.4em}
   \includegraphics[width=0.75\linewidth]{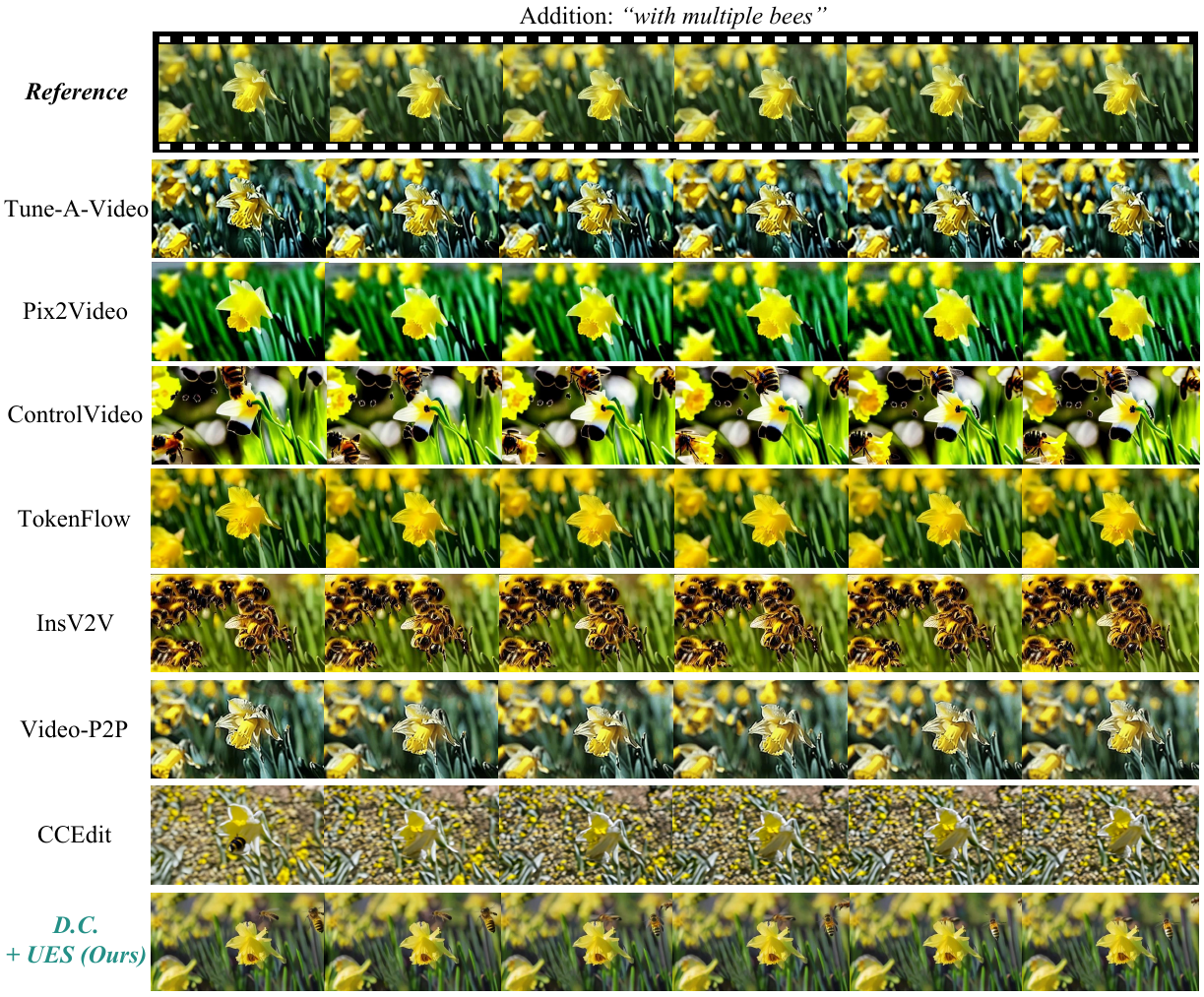}
   \vspace{-0.8em}
    \captionof{figure}{\label{fig:suppl_editing_obj_add}
    Video editing comparison: \textit{Editing-Scenario-Object-Addition}.}
   \vspace{-0.8em}
\end{figure*}

\begin{figure*}[t]
    \centering
    \vspace{-0.8em}
   \includegraphics[width=0.75\linewidth]{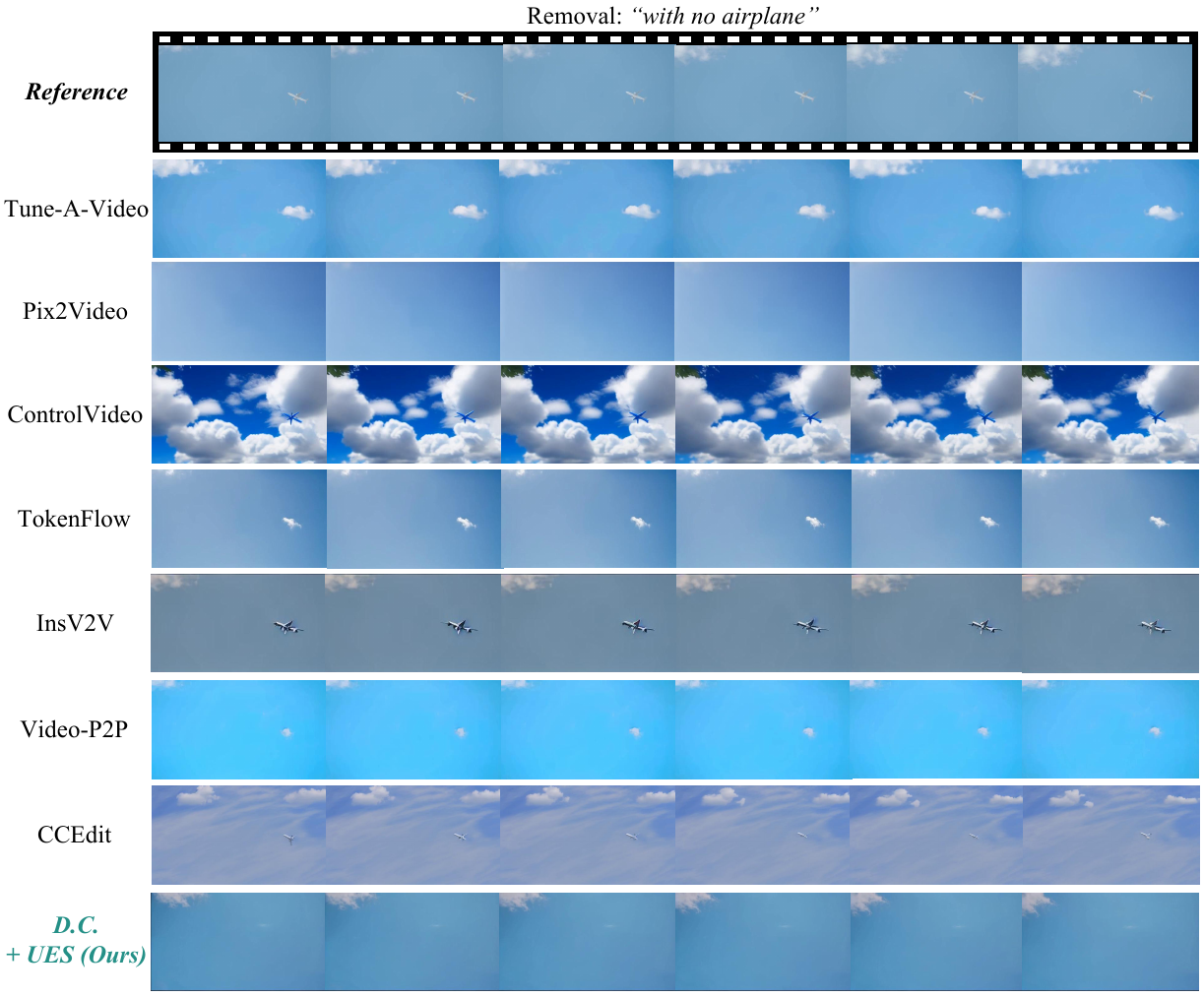}
   \vspace{-0.8em}
    \captionof{figure}{\label{fig:suppl_editing_obj_rem}
    Video editing comparison: \textit{Editing-Scenario-Object-Removal}.}
    \vspace{-0.4em}
\end{figure*}

\begin{figure*}[t]
    \centering
    \vspace{-0.4em}
   \includegraphics[width=0.75\linewidth]{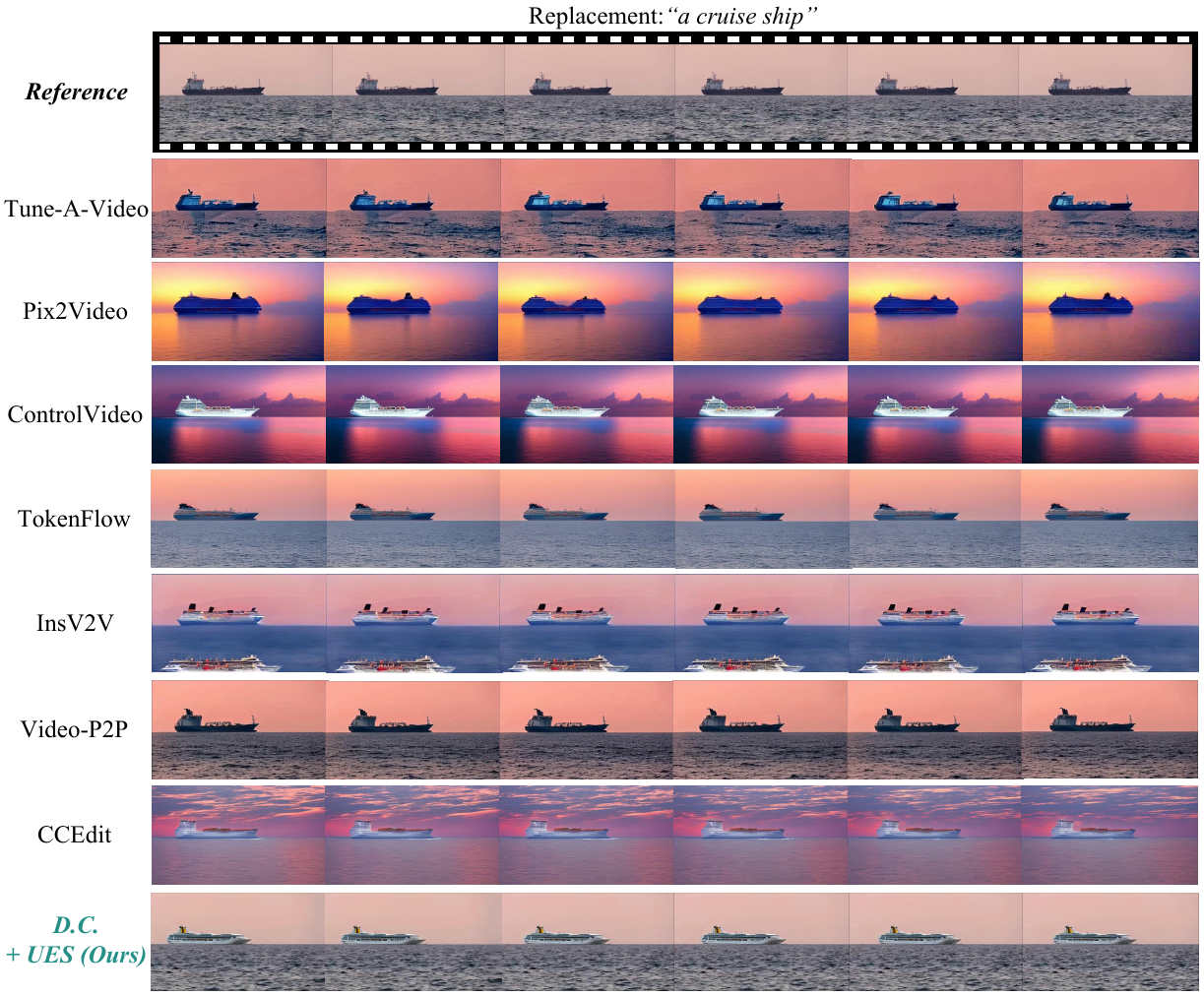}
   \vspace{-0.8em}
    \captionof{figure}{\label{fig:suppl_editing_obj_rep}
    Video editing comparison: \textit{Editing-Scenario-Object-Replacement}.}
    \vspace{-0.8em}
\end{figure*}

\begin{figure*}[t]
    \centering
    \vspace{-0.8em}
   \includegraphics[width=0.75\linewidth]{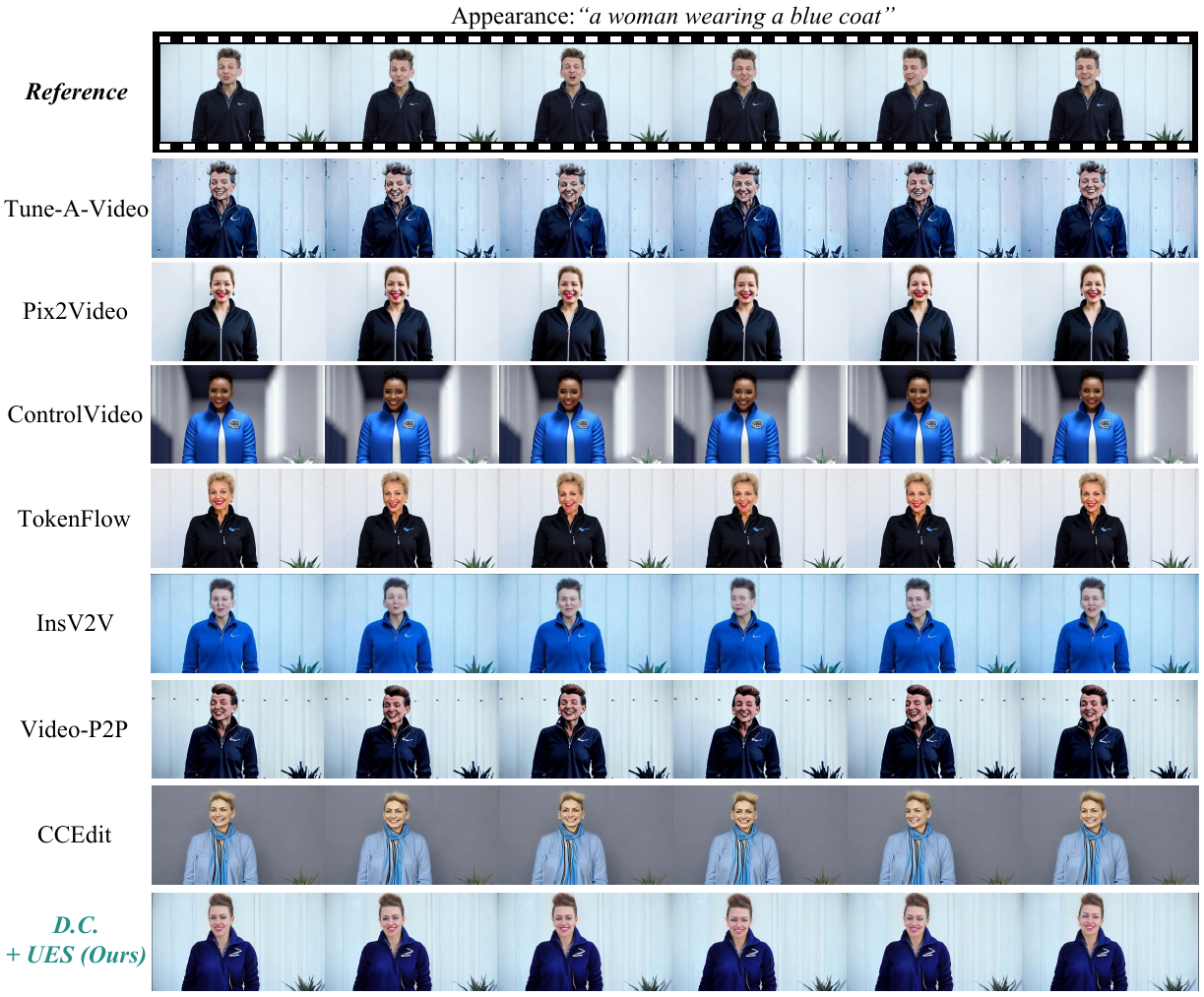}
   \vspace{-0.8em}
    \captionof{figure}{\label{fig:suppl_editing_human_app}
    Video editing comparison: \textit{Editing-Scenario-Human-Appearance}.}
    \vspace{-0.4em}
\end{figure*}

\begin{figure*}[t]
    \centering
    \vspace{-0.4em}
   \includegraphics[width=0.75\linewidth]{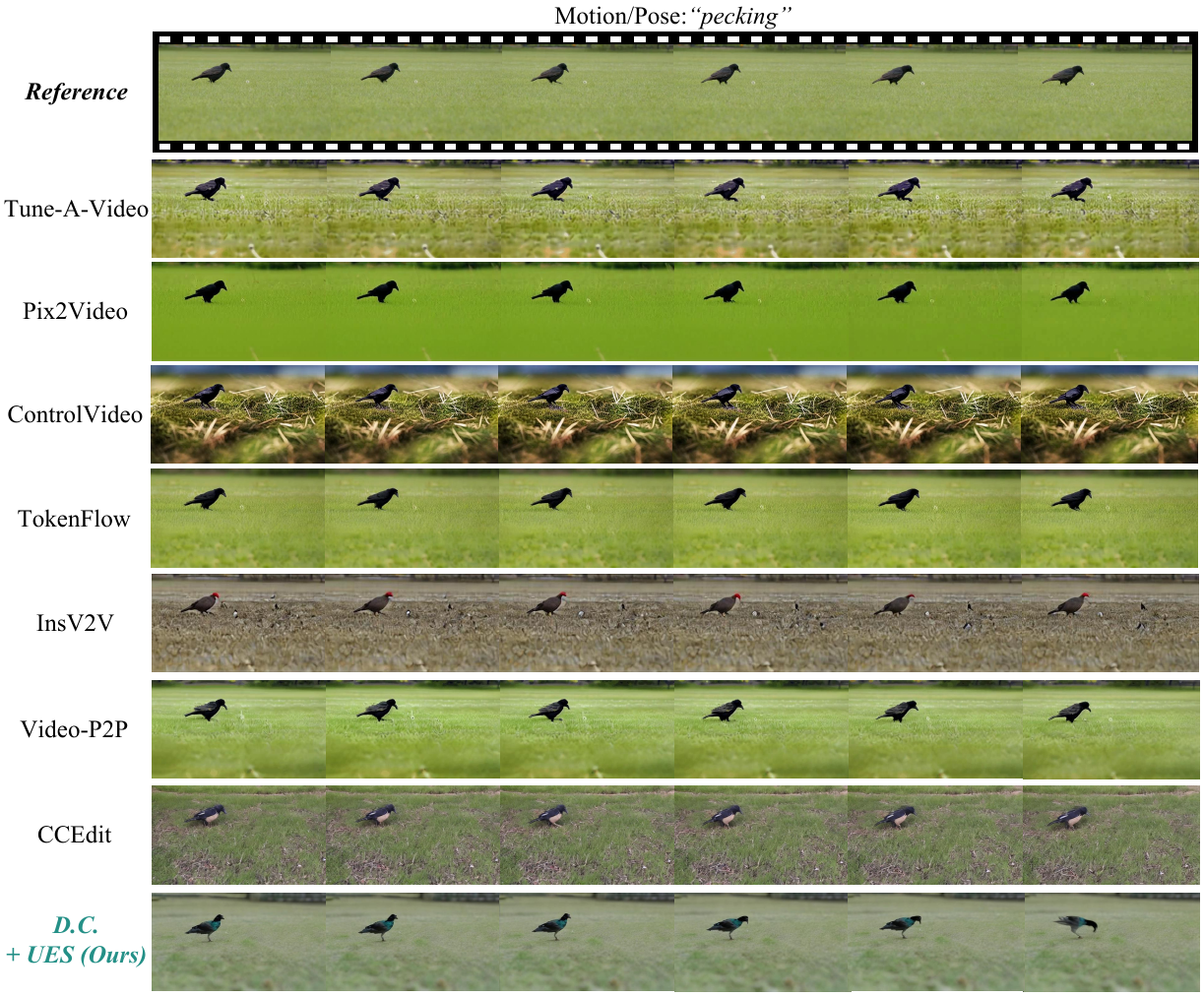}
   \vspace{-0.8em}
    \captionof{figure}{\label{fig:suppl_editing_human_motion}
    Video editing comparison: \textit{Editing-Scenario-Human-Motion}.}
    \vspace{-0.8em}
\end{figure*}

Building on Figure~\ref{fig:quali_video_edit} in the main paper, we present additional comparisons of editing types in Figures~\ref{fig:suppl_editing_type_fore} to \ref{fig:suppl_editing_type_compo}, and editing scenarios in Figures~\ref{fig:suppl_editing_env_weather} to \ref{fig:suppl_editing_human_motion}. Consistent with the main paper's observations, most methods handle the four editing types effectively but exhibit limitations in specific editing scenarios. These findings align with our survey results summarized in Table~\ref{app_tab:review}.

\begin{figure*}[t]
    \centering
    \vspace{-0.4em}
   \includegraphics[width=1\linewidth]{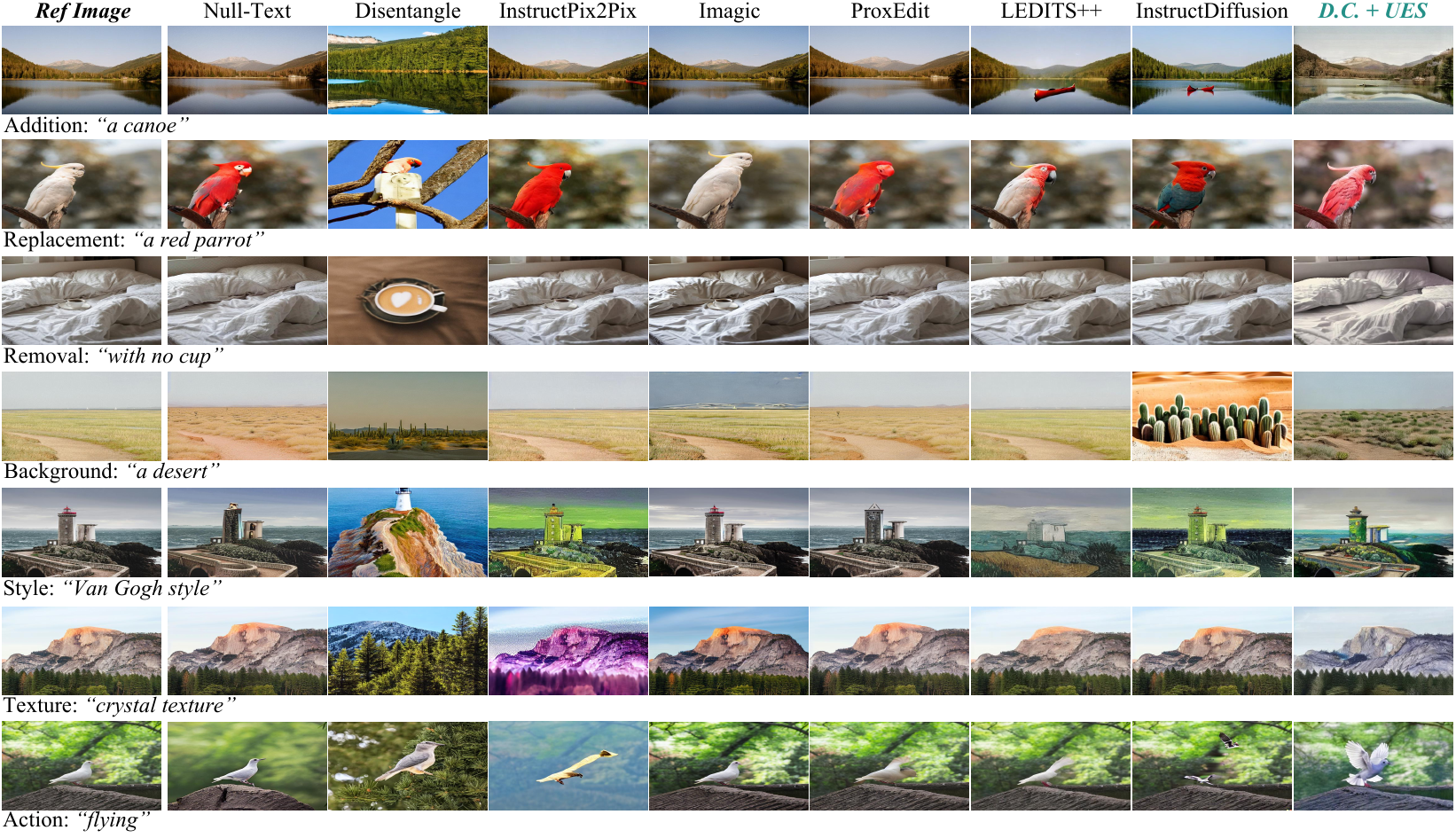}
   \vspace{-2em}
    \captionof{figure}{\label{fig:more_image_edit}
    Image editing comparison.}
    \vspace{-0.4em}
\end{figure*}

\begingroup
\setlength{\tabcolsep}{2.2pt}
\begin{table*}[t]
\renewcommand{\arraystretch}{1.2}
  \centering
  \caption{More comparison with text-guided image editing methods on EditEval \cite{huang2024diffusion}. }
  \centering
  \vspace{-0.8em}
  \scriptsize
   \begin{tabular}{lc|cc|cc|cc|cc|cc|cc|cc}
     \hlineB{2.5}
     \multirow{2}{*}{\textbf{Method}} & \textbf{Additional} & \multicolumn{2}{c}{\textbf{Addition}} & \multicolumn{2}{c}{\textbf{Replacement}} & \multicolumn{2}{c}{\textbf{Removal}} & \multicolumn{2}{c}{\textbf{Background}} & \multicolumn{2}{c}{\textbf{Style}} & \multicolumn{2}{c}{\textbf{Texture}} & \multicolumn{2}{c}{\textbf{Action}}\\
     \cline{3-16}
     & \textbf{Supervision} & \textbf{Mean $\uparrow$} & \textbf{Devia. $\downarrow$} & \textbf{Mean $\uparrow$} & \textbf{Devia. $\downarrow$} & \textbf{Mean $\uparrow$} & \textbf{Devia. $\downarrow$} & \textbf{Mean $\uparrow$} & \textbf{Devia. $\downarrow$} & \textbf{Mean $\uparrow$} & \textbf{Devia. $\downarrow$} & \textbf{Mean $\uparrow$} & \textbf{Devia. $\downarrow$} & \textbf{Mean $\uparrow$} & \textbf{Devia. $\downarrow$} \\
     \hlineB{2}
     Null-Text~\cite{mokady2023null} & \textcolor{red}{\ding{51}} & - & - & 8.15 & 0.80 & -  & - & 7.28 & 1.17 & 7.16 & 0.90 & 6.82 & 1.44 & - & -\\ 
     Disentangle~\cite{wu2023uncovering} & \textcolor{red}{\ding{51}} & 6.14 & 1.74 & 7.66 & 1.41 & -  & - & - & - & - & - & 6.78 & 1.07 & - & -\\
     InstructPix2Pix~\cite{brooks2023instructpix2pix} & \textcolor{red}{\ding{51}} & 6.88 & 2.31 & 5.00 & 1.95 & -  & - & 6.51 & 2.49 & \underline{8.21} & 0.40 & 6.10 & 1.41 & - & -\\ 
     Imagic~\cite{kawar2023imagic} & \textcolor{red}{\ding{51}} & 7.80 & 1.27 & 7.22 & 1.65 & - & - & - & - & 7.26 & 1.25 & 5.57 & 1.49 & 6.97 & 0.80 \\
     ProxEdit~\cite{han2024proxedit} & \textcolor{red}{\ding{51}} & 7.06 & 1.53 & 7.53 & 1.63 & 7.75 & 1.26 & 6.35 & 0.78 & 6.80 & 1.07 & - & - & - & -\\ 
     LEDITS++~\cite{brack2024ledits++} & \textcolor{red}{\ding{51}} & 6.74 & 1.72 & 7.41 & 1.86 & \textbf{8.65}  & 1.29 & 6.91 & 0.97 & 6.86 & 1.20 & - & - & - & -\\ 
     InstructDiffusion~\cite{geng2024instructdiffusion} & \textcolor{red}{\ding{51}} & 7.59 & 1.89 & 6.55 & 1.46 & 7.48 & 1.68 & - & - & 7.41 & 0.66 & \underline{7.13} & 1.83 & - & -\\ 
     \hline
     \rowcolor{lightgray!20}
     VideoCrafter2~\cite{chen2024videocrafter2} & \multicolumn{15}{c}{{\textit{No Editing Capability}}}\\
     \rowcolor{cyan!10}
     $+$ \textbf{\texttt{UES} (Ours)} & \textcolor{green}{\ding{55}} &  \underline{7.62} & 1.60 &  \underline{8.44} & 0.99 & \underline{8.45} & 1.23 & \textbf{7.45} & 0.91 & 8.11 & 0.55 & \textbf{7.20} & 1.24 & \underline{7.48} & 0.97  \\
     \hline
     \rowcolor{lightgray!20}
     DynamiCrafter~\cite{xing2023dynamicrafter} & \multicolumn{15}{c}{{\textit{No Editing Capability}}}\\
     \rowcolor{cyan!10}
     $+$ \textbf{\texttt{UES} (Ours)} & \textcolor{green}{\ding{55}} &  \textbf{7.63} & 1.79 &  \textbf{8.49} & 0.96 & 8.33 & 1.01 & \underline{7.40} & 0.81 & \textbf{8.22} & 0.45 & 6.99 & 1.03 & \textbf{7.53} & 1.11 \\
     \hlineB{2.5}
   \end{tabular}
  \label{tab:quantitative_image_edit}
\end{table*} 
\endgroup

\subsection{Image Editing Evaluation}
\label{app_more_imgedit}

\subsubsection*{More Comparison on EditEval.}
We utilize EditEval \cite{huang2024diffusion}, a comprehensive image editing benchmark covering seven aspects: Addition, Replacement, Removal, Background, Style, Texture, and Action editing. As shown in Table~\ref{tab:quantitative_image_edit}, UES effectively unlocks universal editing capabilities in generation models without requiring additional supervision.

\vspace{-1em}
\subsubsection*{More Image Editing Comparisons.} In addition to Figure~\ref{fig:quali_image_edit} in the main paper, we present a complete comparison of UES on DynamiCrafter in Figure~\ref{fig:more_image_edit} with Null-Text \cite{mokady2023null}, Disentangle \cite{wu2023uncovering}, InstructPix2Pix \cite{brooks2023instructpix2pix}, Imagic \cite{kawar2023imagic}, ProxEdit \cite{han2024proxedit}, LEDITS++ \cite{brack2024ledits++}, and InstructDiffusion \cite{geng2024instructdiffusion}.

\section{Details of User Study}
\label{app_eval_user}

In the context of deep learning \cite{ma2024beyond, chen2024gaussianvton, huang2024crest, chen2024finecliper, yan2024urbanclip}, especially text-guided video editing, it is essential to assess results not only through automated metrics but also through human subjective evaluation. To achieve this, we conducted a comprehensive user study to gather human preferences, comparing our method with recent state-of-the-art models using the mean opinion score (MOS) and direct comparisons.

\begin{figure*}[t]
    \centering
   \includegraphics[width=0.92\linewidth]{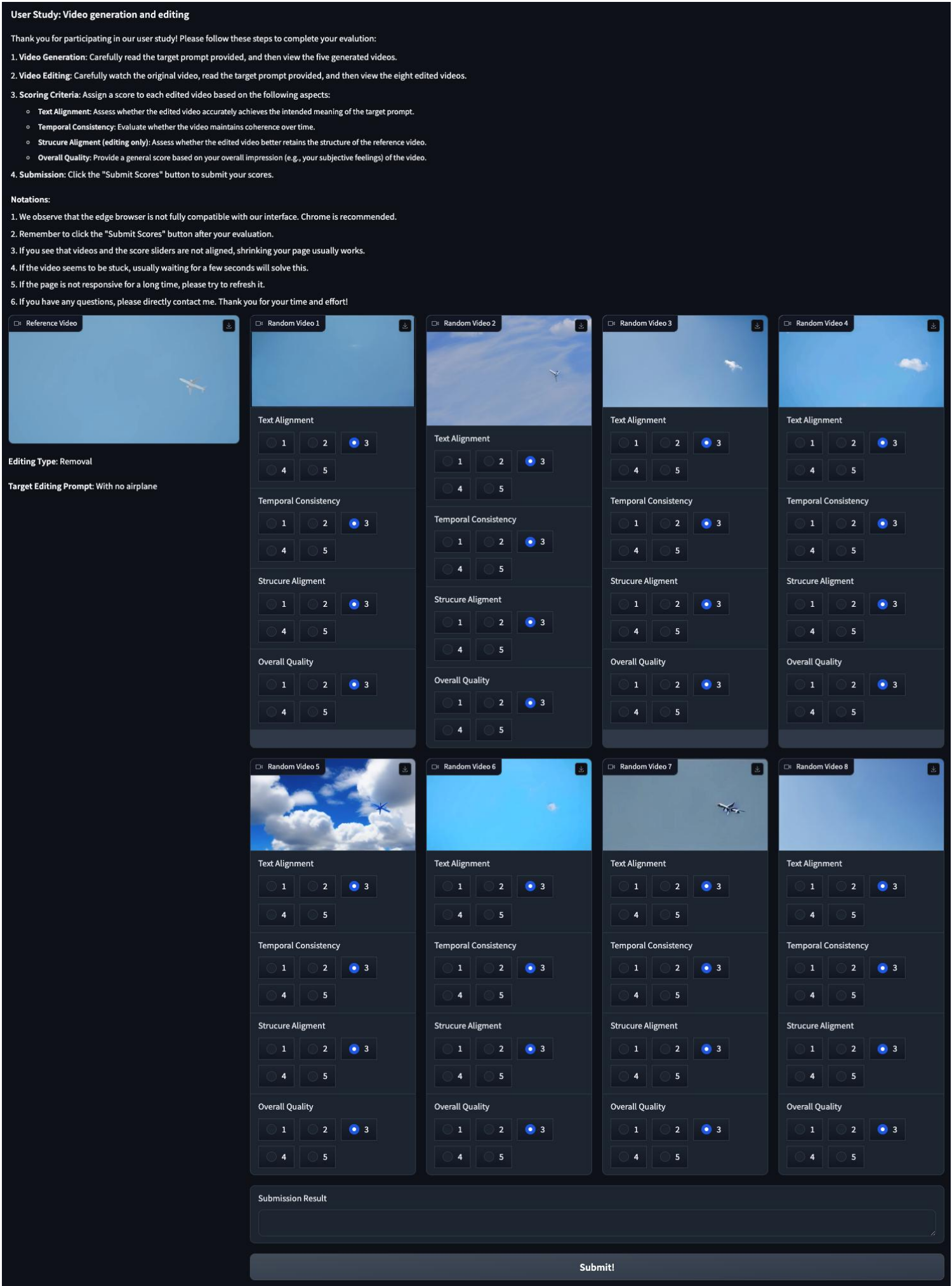}
   \vspace{-0.4em}
    \captionof{figure}{\label{fig:user_study_example}
    Demonstration of our user study interface.}
\end{figure*}

We developed a user-friendly interface and collected scoring results from $15$ volunteers, who evaluated various indicators of the edited videos. Participants rated the videos based on four key aspects: \ding{172} \textit{Text Alignment}: whether the edited video accurately achieves the intended meaning of the target prompt; \ding{173} \textit{Temporal Consistency}: whether the video maintains coherence over time; \ding{174} \textit{Structure Alignment}: whether the edited video better retains the structure of the reference video; and \ding{175} \textit{Overall Quality}: the reflection of the subjective overall rating of the edited video. The interface is demonstrated in Figure~\ref{fig:user_study_example}.

\section{Limitations and Ethical Implications}

\subsubsection*{Limitations.} \textbf{\texttt{UES}} leverages a self-supervised paradigm to unlock universal and superior editing capabilities in generation models, providing an effective approach for unified generation and editing while addressing many challenges in editing models. However, it still has the following limitations:
\begin{itemize}[leftmargin=*]
    \item \textit{Camera Motion Editing}: UES primarily focuses on semantic video content, making it less effective at capturing complex spatial transitions and shot continuity in camera motion, which are difficult to express through text. This can result in unnatural or disjointed camera transitions. Future work could explore using explicit camera motion as an additional condition~\cite{he2024cameractrl}.
    \item \textit{Facial Modeling}: UES relies on high-level semantic information, which may not capture fine-grained facial details such as expressions, textures, or micro-expressions. This can lead to suboptimal facial reconstruction during editing. Incorporating face-specific representations or auxiliary ID control models could enhance facial detail generation~\cite{han2024face}.
    \item \textit{Large Motion Change}: UES struggles with high-speed movements or complex posture shifts due to limitations in static video and text embeddings for capturing temporal dynamics. Future approaches could better utilize motion features from the reference video while avoiding excessive constraints on the original content’s structure, as demonstrated in MOFT~\cite{xiao2024video}.
\end{itemize}

\subsubsection*{Ethical Implications.} UES is developed as a self-supervised fine-tuning strategy for \textit{research only}. It may still raise important ethical considerations, particularly around content manipulation. The ability to generate and edit high-quality videos can potentially be misused for creating misleading or harmful content. To mitigate this risk, we recommend incorporating safeguards such as adding watermarks to edited videos to ensure transparency and authenticity. Additionally, guidelines on responsible use should be established, emphasizing its application in ethical and creative contexts, such as educational, artistic, or research-based scenarios, while discouraging its use for deceptive or harmful purposes.



\end{document}